\documentclass[runningheads]{llncs}

\usepackage{eccv}

\usepackage{eccvabbrv}

\usepackage{graphicx}
\usepackage{booktabs}
\usepackage{marvosym}
\usepackage{booktabs}
\usepackage{makecell}

\usepackage[accsupp]{axessibility}  %

\usepackage{hyperref}

\usepackage{orcidlink}
\usepackage{microtype}
\usepackage{amssymb}
\usepackage{threeparttable}
\usepackage[table]{xcolor}
\definecolor{blue}{RGB}{0, 50, 150}
\usepackage[capitalize]{cleveref}
\crefname{section}{Sec.}{Secs.}
\Crefname{section}{Section}{Sections}
\Crefname{table}{Table}{Tables}
\crefname{table}{Tab.}{Tabs.}

\usepackage{xcolor}
\usepackage{enumitem}
\usepackage{xspace}
\usepackage{booktabs}
\usepackage{colortbl}
\usepackage{multirow}
\usepackage{makecell}
\usepackage{lipsum} %
\usepackage{gensymb} %
\usepackage{wrapfig} %

\newcommand{\Tref}[1]{Table~\ref{#1}}

\newcommand{\fref}[1]{Fig.~\ref{#1}}
\newcommand{\Fref}[1]{Figure~\ref{#1}}

\usepackage[labelsep=period]{caption}
\renewcommand{\paragraph}[1]{\vspace{0.2em}\noindent \textbf{#1 \hspace{0.2em}}}

\definecolor{MyDarkRed}{rgb}{0.66, 0.16, 0.16}
\definecolor{MyDarkBlue}{rgb}{0.16, 0.16, 0.66}

\newcommand{\DatasetName}{AerialMetric\xspace}

\newcommand{\Oblique}{AerialMetric-Oblique\xspace}
\newcommand{\Decoupled}{AerialMetric-Decoupled\xspace}
\newcommand{\Synthetic}{AerialMetric-Synthetic\xspace}
\newcommand{\Wild}{AerialMetric-Wild\xspace}

\begin{document}

\title{AerialMetric: Benchmarking and Adapting \\ UAV Monocular Metric Depth Estimation in the Real World}

\author{
Zhongqiang Song\inst{1} \and
Guanying Chen\inst{1}\thanks{Corresponding author.} \and
Yuqi Zhang\inst{2,3} \and
Yin Zou\inst{1} \and
Chuanyu Fu\inst{1} \and
Zhiyuan Yuan\inst{1} \and
Chuan Huang\inst{4,2} \and
Shuguang Cui\inst{3,2} \and
Xiaochun Cao\inst{1}
}

\authorrunning{Z. Song et al.}
\titlerunning{AerialMetric: UAV Metric Depth Estimation}

\institute{%
$^{1}$Sun Yat-sen University, Shenzhen Campus \quad
$^{2}$FNii--Shenzhen \quad \\
$^{3}$SSE, CUHKSZ  \quad
$^{4}$SIAS, UESTC }

\maketitle
\begin{figure}[h]
    \centering
    \includegraphics[width=\textwidth]{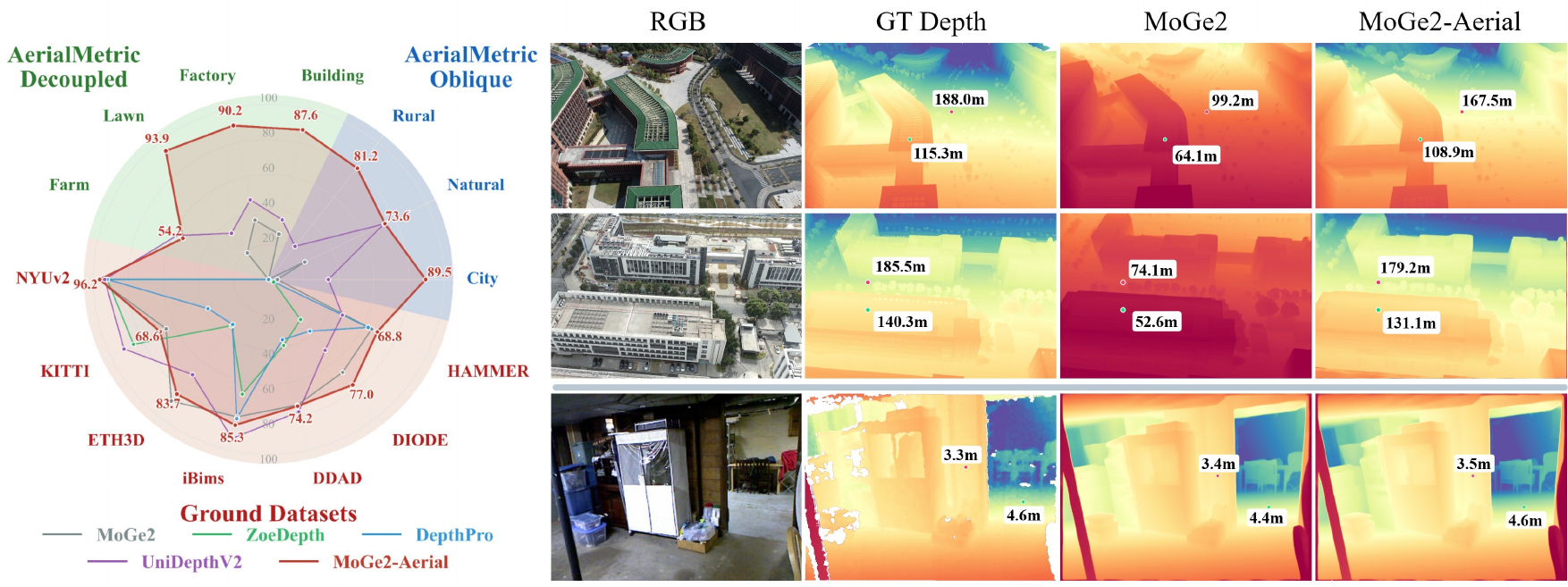}
    \caption{
    By benchmarking existing methods on the AerialMetric dataset, we identify a significant domain gap in aerial depth estimation (the value reported in the left figure is $\delta_{1}$, the higher the better). Fine-tuning the model with our dataset resolves this issue, delivering robust metric depth estimation on both aerial and ground-level scenes. 
    } 
    \label{fig:teaser}

\end{figure}
\begin{abstract}
This paper addresses the problem of monocular metric depth estimation in aerial UAV imagery. Although recent data-driven methods have achieved remarkable progress in ground-level scenarios, models trained primarily on street-view and indoor datasets exhibit significant domain gaps when applied to aerial viewpoints. To tackle these challenges, we introduce \emph{\DatasetName}, a benchmark dataset designed to evaluate and facilitate the adaptation of monocular metric depth estimation under UAV aerial viewpoints. 
The dataset consists of four complementary subsets collected from different sources, jointly covering real-world photogrammetry data, controlled aerial acquisition settings, photorealistic synthetic scenes, and in-the-wild Internet imagery. In total, \emph{\DatasetName} provides 52K real-world and 16K synthetic image–depth pairs with reliable metric ground truth.
Based on this dataset, we conduct systematic evaluations of existing state-of-the-art models under aerial settings and investigate the impact of viewpoint, altitude, and camera parameters on metric depth prediction. 
In addition, by fine-tuning representative metric depth model on our dataset, we establish a comprehensive aerial benchmark and achieve state-of-the-art performance across diverse aerial imagery. Our dataset, code, and model weight are publicly available at
\url{https://kuieless.github.io/AerialMetric-ECCV2026-page/}.
\end{abstract}

\section{Introduction}
\label{sec:intro}

The rapid advancement of aerial robotics technologies has accelerated the large-scale deployment of autonomous Unmanned Aerial Vehicles (UAVs) across diverse applications, \eg, aerial delivery services, infrastructure inspection, environmental monitoring, and public safety.
These emerging applications demand reliable 3D perception of the surrounding environment~\cite{loquercio2021learning,katkuri2024autonomous}. 
Among various perception tasks, monocular metric depth estimation plays a central role by aiming to recover scene geometry with absolute metric scale from a single image.

In recent years, data-driven approaches for depth estimation have achieved remarkable progress, largely driven by the availability of large-scale annotated datasets. 
Deep neural networks trained on diverse depth data have demonstrated strong performance in metric depth prediction settings\cite{li2018megadepth,yin2023metric3d,hu2024metric3d,wang2025moge,wang2026moge
,yang2024depth,yang2024depth2,lin2025depth3}.
However, despite their strong performance, state-of-the-art metric depth estimation models suffer from a significant domain gap between ground-level data and aerial UAV imagery, resulting in limited generalization to aerial viewpoints, where significantly different viewing geometries and bird's-eye perspectives dominate.

Currently, there is a lack of real-world, diverse, and variable-controlled aerial datasets with metric depth,
which limits effective model adaptation and prevents reliable learning of absolute scene scale. 
In addition, the absence of an evaluation benchmark makes it difficult to systematically assess existing methods under diverse UAV imaging conditions. 
Despite rapid advances in depth estimation architectures, aerial metric depth estimation remains substantially behind its ground-level counterpart (see~\fref{fig:teaser}).

\begin{figure*}[tb] \centering
    \includegraphics[width=\textwidth]{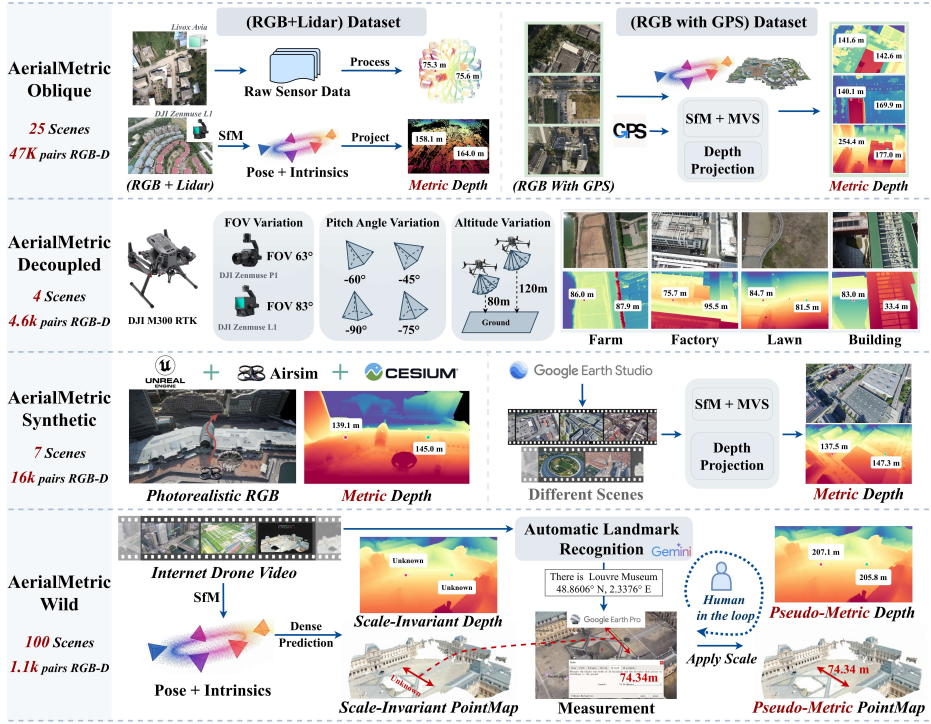}
    \caption{
    Data collection pipeline of the proposed \DatasetName dataset. } \label{fig:pipeline}
\end{figure*}

To bridge this gap, we introduce \textbf{\DatasetName}, a large-scale benchmark for monocular metric depth estimation in aerial scenarios. 
Our dataset consists of four complementary components, as shown in \fref{fig:pipeline}.
\emph{\Oblique} is constructed from publicly available real-world oblique aerial imagery, from which we process and curate high-quality image-depth pairs with reliable metric ground truth, comprising 47K samples for training and evaluation.
\emph{\Decoupled} is collected by ourselves through systematic photogrammetry acquisition, where each scene is captured under controlled variations of viewing angles, fields of view, and flight altitudes (four angles, two FOVs, and two heights). This subset enables a disentangled analysis of how aerial imaging configurations affect metric depth estimation.
\emph{\Synthetic} contains 16K photorealistic image-depth pairs for training rendered using Unreal Engine and Google Earth, providing controlled, multi-angle views of scenarios that are difficult to capture in real-world scenes.
\emph{\Wild} consists of diverse in-the-wild aerial images collected from the Internet, serving as a challenging evaluation set for cross-domain generalization.

Based on the proposed \emph{\DatasetName} dataset, we conduct a comprehensive analysis of existing metric depth estimation methods under aerial settings. 
Our study reveals that current state-of-the-art models exhibit limited robustness and degraded performance when applied to aerial imagery.
In addition, leveraging the high-quality metric depth supervision provided by \Oblique and \Synthetic, we fine-tune a representative metric depth estimation model, which achieves enhanced performance under this challenging aerial setting, demonstrating that our dataset effectively improves the generalization capability of monocular metric depth estimation methods.

In summary, the key contributions of this paper are as follows:
\begin{itemize}
    \item We introduce \DatasetName, an aerial monocular metric depth dataset comprising four complementary subsets, significantly expanding the scale, diversity, and geometric coverage of aerial depth data.

    \item We establish a comprehensive benchmark for monocular aerial metric depth estimation, systematically analyzing the impacts of viewing angle, flight altitude, and camera field-of-view.

    \item By adapting representative state-of-the-art models on \DatasetName dataset, we achieve state-of-the-art performance across diverse UAV deployment scenarios, demonstrating that our dataset effectively improves the generalization capability of monocular metric depth estimation methods.
\end{itemize}

\section{Related Work}
\label{sec:related_works}

\paragraph{Monocular Depth Estimation}
Monocular depth estimation has advanced significantly with deep learning and large-scale visual models. 
Early deep learning-based methods \cite{eigen2014depth} focused on multi-scale feature learning. 
DPT \cite{Ranftl2021} leverages Vision Transformers to advance monocular depth estimation.
The MiDaS series \cite{Ranftl2022, birkl2023midas} enables robust zero-shot relative depth estimation across scenes using large mixed datasets. The Depth Anything series \cite{yang2024depth, yang2024depth2, lin2025depth3} enhances zero-shot generalization with massive unlabeled data and cross-view consistency. The Lotus series \cite{he2024lotus, he2025lotus2} leverages diffusion priors to achieve strong generalization and high-precision geometry estimation.

Metric depth is essential for grounded 3D tasks. To unify relative and metric predictions, ZoeDepth \cite{bhat2023zoedepth} and Metric3Dv2 \cite{yin2023metric3d, hu2024metric3d} achieve zero-shot metric depth estimation. MoGe2 \cite{wang2026moge, wang2025moge} and MapAnything \cite{keetha2026mapanything} improve open-domain metric reconstruction. InfiniDepth \cite{yu2026infinidepth} uses neural implicit fields for arbitrary-resolution depth prediction. Despite the remarkable success in terrestrial scenes, existing methods still struggle when deployed in UAV scenes.

Compared to depth estimation from ground perspectives, UAV-based monocular depth estimation \cite{yu2023scene,shimada2022pix2pix,florea2025tandepth,madhuanand2020deep} remains challenging, due to the scarcity of dedicated, large-scale, variable-controlled, real-world aerial depth datasets.

\paragraph{General Datasets with Depth}
The development of monocular depth estimation is deeply rooted in the availability of datasets.
NYUv2 \cite{2012nyu} and ScanNet \cite{dai2017scannet} provide abundant structured data captured by RGB-D sensors, laying the foundation for diverse 3D tasks.
KITTI \cite{geiger2013vision} and Waymo \cite{2020waymo} are built for autonomous driving scenarios, equipped with LiDAR data. nuScenes \cite{2020nuScenes} further supplements millimeter-wave radar to achieve full-view coverage, improving outdoor generalization. Argoverse \cite{chang2019argoverse,wilson2023argoverse} provides depth supervision through LiDAR point clouds and stereo image pairs. DAP \cite{lin2025dap} extends the depth estimation task to panoramic views, enabling zero-shot generalization in panoramic scenes. Although these general datasets have advanced monocular depth estimation, they are almost limited to ground-level perspectives.

\paragraph{UAV Datasets}
UAV data presents challenges for visual perception due to its unique viewpoints and wide field-of-view. Existing UAV datasets like VisDrone \cite{zhu2021detection} and UAVid \cite{2020UAVid} focus on detection \cite{Detection1,Detection2}, tracking \cite{Tracking1,Tracking2}, and segmentation \cite{Segmentation1,Segmentation2}, but lack 3D geometric ground truth.
UAVLight \cite{du2026uavlight} focuses on illumination variations, and OpenFly \cite{OpenFly} focuses on vision-and-language navigation. Neither of them directly supports high-precision 3D tasks.

To support a broader range of 3D perception tasks \cite{yu2023scene,2022meganerf,zhang2024aerial}, UAV datasets have expanded into two complementary directions: geometry-centric benchmarks for semantic 3D understanding and depth perception, such as STPLS3D \cite{chen2022stpls3d}, DDOS \cite{kolbeinsson2024ddos}, and FIReStereo \cite{dhrafani2025firestereo}; and localization/navigation benchmarks, including NTU VIRAL \cite{nguyen2022ntu}, CrossLoc \cite{yan2022crossloc}, GraCo \cite{zhu2023graco}, MUN-FRL \cite{thalagala2024mun}, and UAVD4L \cite{wu2024uavd4l}.

OpenDroneMap \cite{opendronemap} provides an open-source photogrammetry platform that facilitates large-scale aerial surveying and mapping.
Mid-Air \cite{2019MidAir} and SynDrone \cite{rizzoli2023SynDrone} have constructed UAV datasets via synthetic data generation to be applicable to depth estimation tasks. ClaraVid \cite{beche2025claravid} provides a synthetic aerial benchmark with dense depth and semantic annotations.

Recent UAV benchmarks have evolved from synthetic data toward real-world scenes. 
UseGeo \cite{nex2024usegeo}, WildUAV \cite{florea2021wilduav}, and UAVid-3D-Scenes \cite{florea2025tandepth} provide real UAV imagery with LiDAR- or photogrammetry-derived depth supervision.
DublinCity \cite{2019dublincity}, UrbanBIS \cite{UrbanBIS}, H3D \cite{KOLLE2021H3D}, and GauU-Scene \cite{xiong2024gauuscene,xiong2024gauuscenev2assessingreliability} provide urban UAV datasets with point clouds or LiDAR ground truth and semantic annotations. MARS-LVIG \cite{2024MARSLVIG} and UAVScenes \cite{wang2025uavscenes} offer multi-sensor and semantically annotated data. UrbanScene3D \cite{UrbanScene3D} and FlyAwareV2 \cite{BARBATO2026FlyAwareV2} build real-synthetic hybrid benchmarks for outdoor 3D modeling. AerialMegaDepth \cite{vuong2025aerialmegadepth} addresses aerial-ground viewpoint differences, and OccuFly \cite{gross2026occufly} supports semantic scene completion with metric depth and occupancy annotations.

However, existing UAV benchmarks are constrained by inconsistent data modalities, synthetic-to-real domain gaps, and limited flight scenarios, hindering unified 3D perception. We introduce \emph{AerialMetric}, a large-scale aerial depth benchmark designed to overcome these limitations and address scale ambiguity for UAV 3D sensing (see~\Tref{tab:dataset_comparison}).

\begin{table*}[t]
\centering
\caption{Comparison between existing datasets and our proposed AerialMetric dataset. Our dataset comprises four complementary subsets designed to cover diverse scenarios, including real-world photogrammetry, controlled aerial acquisitions with decoupled flight parameters, photorealistic synthetic scenes, and in-the-wild imagery.
}
\label{tab:dataset_comparison}
\resizebox{\textwidth}{!}{
\setlength{\tabcolsep}{5pt}
\Large
\begin{tabular}{l c c c c c c c c c c}
\toprule
\multirow{2}{*}{\textbf{Dataset}} & \multirow{2}{*}{\textbf{Type}} & \multirow{2}{*}{\textbf{Raw Data}} & \multirow{2}{*}{\textbf{Depth GT}} & \multirow{2}{*}{\textbf{3D Assets}} & \multirow{2}{*}{\textbf{Area ($\text{m}^2$)}} & \multirow{2}{*}{\textbf{Scenes}} & \multirow{2}{*}{\textbf{\#Images}} & \multicolumn{3}{c}{\textbf{Flight Parameters (Decoupled)}} \\
\cmidrule(lr){9-11}
& & & & & & & & \textbf{Pitch} & \textbf{Altitude} & \textbf{FOV} \\
\midrule
University-1652 \cite{zheng2020university1652}
    & Synthetic & RGB & None & - & - & 1652 & $5.0 \times 10^4$ & - & - &- \\
UAVLight \cite{du2026uavlight}
    & Real & RGB & None & Point Cloud+Mesh & >$6.1 \times 10^5$  & 18 & 4385 & Oblique & - & - \\
Mill 19 (Mega-NeRF \cite{2022meganerf})  
    & Real & RGB & None & -  & >$1.3 \times 10^5$ & 2 & >$3.6 \times 10^3$ & - & - & - \\
VisDrone \cite{zhu2021detection}
    & Real & RGB & None & - & - & 14 & >$2.7 \times 10^5$ & - & - & - \\
Mid-Air \cite{2019MidAir}
    & Synthetic & RGB-D & Metric & - & - & 2  & >$4.2 \times 10^5$& - & - & - \\
DublinCity \cite{2019dublincity}
    & Real & RGB+LiDAR & Sparse Lidar & Point Cloud & >$2.0 \times 10^6$ & 1 & 8504 & Oblique & - & - \\
Hessigheim \cite{KOLLE2021H3D}
    & Real & RGB+LiDAR & Sparse Lidar & Point Cloud+Mesh & >$1.9 \times 10^5$ & 1 & - & Oblique & - & - \\
MegaDepth \cite{li2018megadepth}
    & Real & RGB-D & Relative & Point Cloud & - & 196 & >$1.3 \times 10^5$ & - & - & - \\
MARS-LVIG \cite{2024MARSLVIG}
    & Real & RGB+LiDAR & Sparse LiDAR & Point Cloud & - & 4 & - & - & - & - \\
UAVScenes \cite{wang2025uavscenes}
    & Real & RGB+LiDAR & Sparse LiDAR & Point Cloud+Mesh & - & 21 & $1.2 \times 10^5$ & - & - & - \\
OpenDroneMap \cite{opendronemap} 
    & Real & RGB & None & Point Cloud+Mesh & - & 41 & 15061 & Oblique & - & - \\
Esri Sample Drone Datasets \cite{esri} 
    & Real & RGB+LiDAR & Sparse LiDAR & Point Cloud+Mesh & - & 8 & - & Oblique & - & - \\
UrbanBIS \cite{UrbanBIS} 
    & Real & RGB & None & Point Cloud+Mesh & $1.1 \times 10^7$ & 6 & $1.1 \times 10^5$ & Oblique & - & - \\
UrbanScene3D \cite{UrbanScene3D}
    & Real+Synthetic & RGB+LiDAR & Sparse Lidar & Point Cloud+Mesh & $1.36\times10^8$ & 16 & >$1.3 \times 10^5$ & - & - & - \\
GauU-Scene V2 \cite{xiong2024gauuscenev2assessingreliability}
    & Real & RGB+LiDAR & Sparse LiDAR & Point Cloud & $6.7 \times 10^6$ & 6 & 4693 & - & - & - \\
OccuFly \cite{gross2026occufly}
    & Real & RGB-D & Metric & Point Cloud & >$1.9 \times 10^5$  & 9 & >$2.0 \times 10^4$ & - & 3 Levels & - \\
\midrule
\textbf{\Oblique}
    & Real & RGB/RGB+LiDAR & Metric & Point Cloud+Mesh & $6.7 \times 10^6$ & 25 & >$4.7 \times 10^4$ & Oblique & - & - \\
\textbf{\Decoupled}
    & Real & RGB & Metric & Point Cloud+Mesh & $2.9 \times 10^5$ & 4 & >4600 & 4 Angles & 2 Levels & 2 Settings \\
\textbf{\Synthetic}
    & Synthetic & RGB-D & Metric & Mesh Partly & $2.4 \times 10^7$ & 7  & >$1.6 \times 10^4$ & 4 Angles & 3 Levels & 3 Settings \\
\textbf{\Wild}
    & Real & RGB & Pseudo-Metric & Point Cloud & - & 100  & >1100 & - & - & - \\
\bottomrule
\end{tabular}
}

\end{table*}

\section{The AerialMetric Dataset}
\begin{figure*}[tb] \centering
    \includegraphics[width=\textwidth]{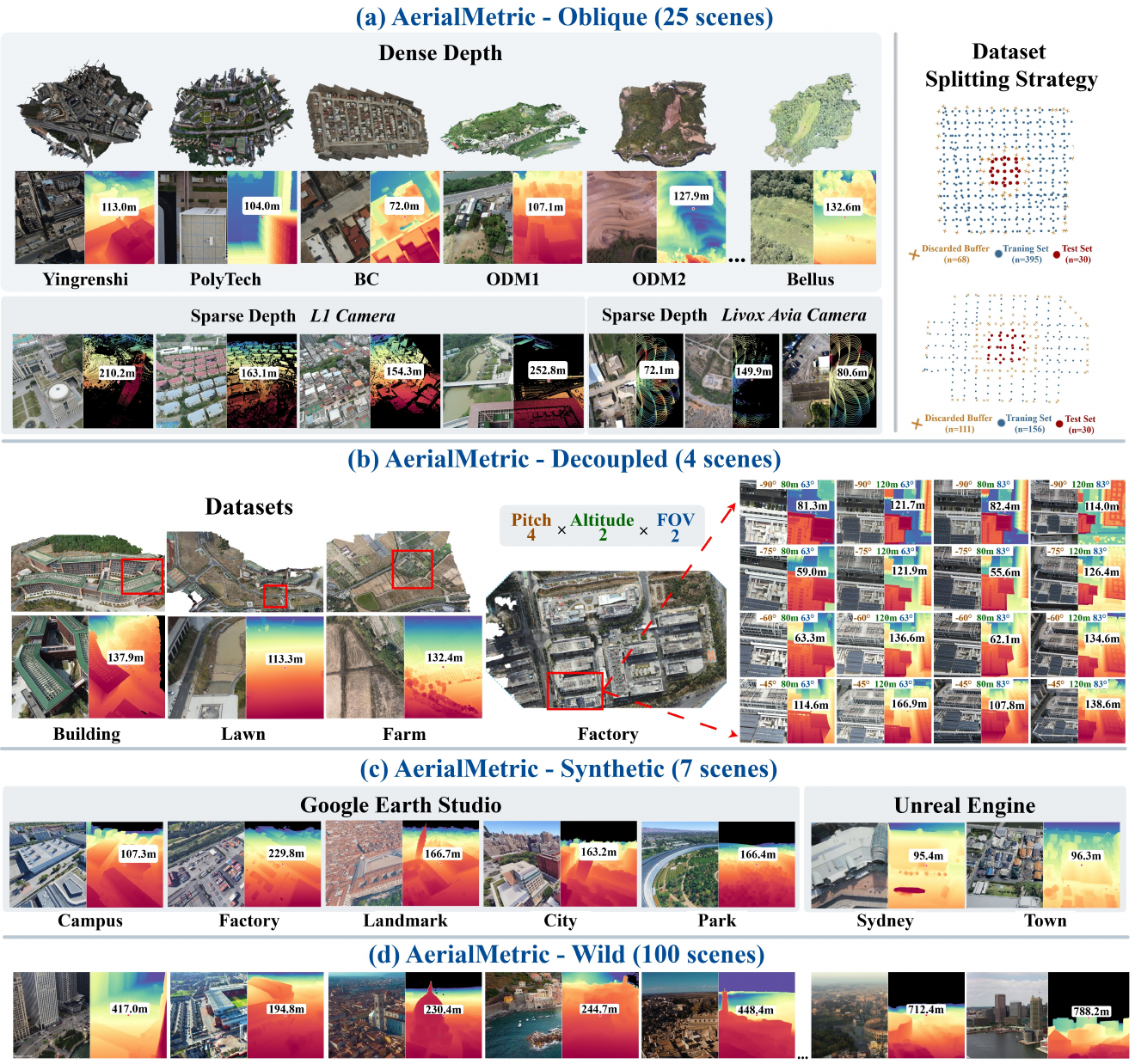}
    \caption{
    Visualization of the introduced \DatasetName dataset.} \label{fig:dataset}
\end{figure*}
\paragraph{Overview} Existing metric depth datasets remain largely constrained to horizon-level views and therefore fail to cover the drastic altitude and pitch variations, and wide depth ranges inherent in aerial imagery. 
To address this distribution shift, we introduce \emph{AerialMetric}, a large-scale, high-quality aerial depth dataset built through a hybrid construction pipeline that achieves geometric accuracy, parameter decoupling, and scene generalization. 

The dataset comprises four complementary components (see \fref{fig:dataset}): 
(1) \emph{AerialMetric-Oblique}, an oblique photography dataset curated from publicly available photogrammetry sources, featuring high-quality GPS-positioned imagery (a subset of which includes associated LiDAR data);
(2) \emph{AerialMetric-Decoupled}, an RTK-equipped UAV benchmark performing rigorous variable-decoupled sampling across altitudes, pitch angles, and fields of view to enable quantitative evaluation of capture variations; 
(3) \emph{AerialMetric-Synthetic}, synthetic data generated by utilizing Google Earth Studio and simulation engines (UE + AirSim) to cover long-tail scenarios involving extreme viewpoints and complex lighting conditions, while enabling decoupled capture variations;
and (4) \emph{AerialMetric-Wild}, which consists of annotated internet drone videos with metric depth maps for in-the-wild evaluation through a human-in-the-loop scale calibration pipeline that leverages known landmark dimensions to resolve monocular scale ambiguity.

\subsection{\Oblique: UAV Photogrammetry Datasets}
\paragraph{Data Sources and Aggregation}
To construct a high-fidelity aerial depth benchmark, \Oblique aggregates and standardizes six prominent urban-scale photogrammetry datasets: \emph{UrbanBIS} \cite{UrbanBIS}, \emph{GauU-SceneV2} \cite{xiong2024gauuscenev2assessingreliability}, \emph{UAV-Scenes} \cite{wang2025uavscenes}, \emph{UrbanScene3D} \cite{UrbanScene3D}, \emph{ODM} \cite{opendronemap}, and \emph{ESRI} \cite{esri}. We construct over 47K image-depth pairs from 25 sub-datasets, spanning diverse scene categories: \textit{City}, \textit{Rural}, and \textit{Natural}. To mitigate high inter-frame redundancy in the \emph{UAVScenes} subset, we apply a 5-frame downsampling interval following the authors' practice.

To accommodate heterogeneous sensor modalities, we design a processing pipeline that categorizes depth recovery into active LiDAR and passive photogrammetry paradigms.

\paragraph{Ground-truth Depth from LiDAR} 
For the \emph{UAVScenes} and \emph{GauU-SceneV2} datasets, we process the raw point clouds captured by LiDAR into per-frame metric depth maps. \emph{UAVScenes} provides single-frame point clouds, enabling direct projection onto the image plane. 
Since \emph{GauU-SceneV2} does not provide per-frame LiDAR data parsing functionality, we use the professional commercial software DJI Terra~\cite{DJITERRA} to generate a unified point cloud and project it onto individual image views.
A critical challenge in LiDAR projection is ray penetration artifacts, where rays from background points erroneously pass through sparse foregrounds and manifest as noise in the resulting depth maps.
To address this, we reconstruct a continuous mesh from the data and validate projected depths against this mesh surface. Points exhibiting a discrepancy greater than 1.0\,m relative to the mesh are discarded as occlusion noise.

\paragraph{Ground-truth Depth from Photogrammetry} 
For \emph{UrbanBIS}, \emph{UrbanScene3D}, \emph{ODM}, and \emph{ESRI}, we utilize the RGB images and reconstruct 3D meshes via DJI Terra at the highest native resolution to preserve high-frequency details that would otherwise be lost during downsampling.
Metric depth maps are then derived via pose-guided mesh rasterization, ensuring pixel-level correspondence between RGB images and depth values.

\paragraph{Dataset Splits.} To enable rigorous zero-shot evaluation, we implement a strict train-test split that prevents geographic data leakage. As illustrated in \fref{fig:dataset}~(a), we exclude the outermost cameras and prune over 4.4K frames within a 50\,m buffer zone of the test regions. The final test set comprises 1.4K images curated from over 45 distinct geographical regions across the 25 sub-datasets. The remaining data is allocated for training. Detailed statistics and split protocols are provided in the supplementary material.

\subsection{\Decoupled: Variable Decoupled Dataset}

To systematically evaluate the impact of imaging geometry on monocular depth estimation, we construct \Decoupled, a strictly controlled, variable-decoupled benchmark. Unlike conventional datasets that entangle multiple confounding factors, 

\Decoupled disentangles the effects of camera pose (pitch angle, flight altitude) and field of view (FOV) via orthogonal sampling.
As detailed in Table~\ref{tab:am_bench_params}, we utilize an RTK-equipped UAV platform to sample across four distinct scenes. This dataset is designed exclusively for testing, providing a controlled evaluation protocol free from training bias.

\paragraph{Acquisition Platform}
Data acquisition is conducted using a DJI Matrice 300 RTK platform equipped with dual payloads: (1) Zenmuse L1 (LiDAR/RGB) and (2) Zenmuse P1 (Full-frame RGB). The RTK system provides centimeter-level absolute spatial alignment, ensuring precise control over camera positioning and enabling accurate metric depth ground truth. The captured images feature resolutions of $8192 \times 5460$ (P1 camera) and $5472 \times 3648$ (L1 camera).

\paragraph{Variable-Decoupled Acquisition Protocol} 

We strictly decouple four core variables through orthogonal sampling:
(1) \emph{Pitch Angle}: varying from nadir to oblique perspectives ($\theta \in \{-90^\circ, -75^\circ, -60^\circ, -45^\circ\}$); 
(2) \emph{Relative Altitude}: captured at $80$m and $120$m above ground level to assess scale sensitivity; 
(3) \emph{FOV}: leveraging the distinct sensor characteristics (L1 at $83^\circ$ HFOV, P1 at $63^\circ$ HFOV); 
and (4) \emph{Scene Types}: encompassing four distinct semantic environments: \emph{Building}, \emph{Lawn}, \emph{Farmland}, and \emph{Industrial Factory}.
\begin{wraptable}{r}{0.45\textwidth}
    \centering
    \caption{UAV acquisition parameters for the \Decoupled dataset.}
    \label{tab:am_bench_params}
    \resizebox{\linewidth}{!}{%
\begin{tabular}{@{} l l @{}}
\toprule
\textbf{Parameter} & \textbf{Specification} \\
\midrule
Positioning System & \makecell[l]{\textbf{Real-Time Kinematic (RTK)} \\ (Centimeter-level accuracy)} \\
UAV Platform       & DJI Matrice 300 RTK \\
\midrule
Sensor \& FOV      & \makecell[l]{L1: $83^\circ$ (LiDAR Multi-modal Payload) \\ P1: $63^\circ$ (Full-frame RGB Payload)} \\
Relative Altitude  & 80 m, 120 m (Dual-scale assessment) \\
Gimbal Pitch       & $-90^\circ$ (Nadir), $-75^\circ$, $-60^\circ$, $-45^\circ$ (Oblique) \\
\midrule
Image Resolution   & \makecell[l]{L1: $5472 \times 3648$; \quad P1: $8192 \times 5460$} \\
Flight Speed       & 10 m/s \\
Trigger Interval   & Spatial equidistant, 10 m/frame \\
Exposure Strategy  & Automatic \\
\midrule
Scene Category   &
    \begin{tabular}{@{} p{1.5cm} p{1.5cm} p{1.5cm} p{1.5cm} @{}}
    \emph{Building} & \emph{Lawn} & \emph{Farm} & \emph{Factory}
    \end{tabular} \\
    Area ($\times 10^4$ m$^2$) &
    \begin{tabular}{@{} p{1.5cm} p{1.5cm} p{1.5cm} p{1.5cm} @{}}
    2.6 & 7.2 & 9.0 & 10.0
    \end{tabular} \\
\bottomrule
\end{tabular}%
}

\end{wraptable}
For each scene, we captured 16 flight trajectories with strictly controlled combinations of FOV, pitch, and altitude variables, yielding approximately 4{,}600 image-depth pairs.

\paragraph{Ground Truth Curation}
To mitigate geometric degradation at reconstruction boundaries, we extend flight trajectories with high-overlap oblique captures surrounding the core test zones. We employ DJI Terra for 3D reconstruction, processing the captured images to generate high-fidelity geometric models. Depth maps are derived only from the central regions through projection onto the reconstructed geometry, discarding boundary artifacts. Additionally, dynamic entities are manually annotated and explicitly excluded from evaluation to ensure static scene consistency.

\subsection{\Synthetic: Variable Controlled Rendering}
To overcome the physical constraints of real-world UAV collection, we construct a two-component synthetic dataset for training data augmentation.

\paragraph{Google Earth Studio Dataset}
We render multi-view sequences in Google Earth Studio (GES) across five diverse scenarios (City, Campus, Factory, Landmark, and Park) for training. We employ a rigorously decoupled parameter grid: three altitude bins in (70-300m), four pitch angles from -90° to -45°, and three FOVs, producing $2{,}000$ RGB images at $3840\times2160$ resolution. Since GES lacks native depth export, 
we reconstruct 3D meshes using DJI Terra and project metric depth via pose-guided rasterization. 

\paragraph{Unreal Engine Dataset}
To efficiently scale training data, we integrate Unreal Engine, AirSim, and Cesium to stream Google Earth 3D Tiles. This simulator directly extracts native Z-buffers, by passing photogrammetric reconstruction. The framework generates extreme trajectories (e.g., low-to-medium-altitude UAV flights, acute side-views) with randomized altitudes ($80$--$200$\,m), pitches, and FOVs. The pipeline produces approximately 16K RGB-D pairs at $3840\times2160$ resolution, broadening spatial distribution and improving generalization to out-of-distribution wild viewpoints.

\subsection{\Wild: Reliable Metric Pseudo-Ground Truth}

Existing metric depth datasets lack open-domain diversity due to their reliance on specialized sensors and controlled acquisition environments. To address this limitation, we construct \Wild by exploiting unlabeled Internet drone videos through a human-in-the-loop \textit{Scale Recovery Framework}. This dataset is designed exclusively for in-the-wild evaluation, featuring authentic drone trajectories, diverse FOVs, and unconstrained viewing conditions. 
 
\paragraph{Multi-View Relative Depth Estimation}
Instead of relying on single-image estimation, we explicitly leverage the temporal multi-view nature inherent to drone tour sequences. We employ COLMAP~\cite{schonberger2016structure} for initial camera pose estimation and Depth Anything 3 (DA3)~\cite{lin2025depth3} for dense depth prediction. This sequence-level conditioning yields a shape-reliable but unscaled relative point cloud $\mathcal{P}_{rel}$, capturing geometric structure without metric constraints.

\paragraph{Human-in-the-loop Metric Scale Calibration}
Monocular reconstruction recovers geometry only up to an unknown scale. To obtain metric depth, we calibrate each sequence using a small set of human-annotated landmark pairs. Annotators select semantically clear point pairs in an interactive interface, forming a constraint set $\mathcal{K}$. For each pair, we measure its distance in the unscaled point cloud $\mathcal{P}_{rel}$ and assign a real-world distance from geographic tools (\ie, Google Earth Pro).
We then estimate a global scale factor by averaging the ratio of real-world distance to reconstructed distance across all selected pairs, i.e., $s=\mathrm{avg}_k\!\left(D_{gt}^{(k)}/d_{rel}^{(k)}\right)$. Applying this single factor to all points maps $\mathcal{P}_{rel}$ into metric space and yields $\mathcal{P}_{metric}$.

Since manual selection, reference measurements, and depth inference are all imperfect, $\mathcal{P}_{metric}$ should be treated as pseudo-metric supervision. Despite these limitations, the dataset remains valuable for evaluating method performance in in-the-wild scenarios.

\paragraph{Dataset Composition}
We curated 100 high-fidelity sequences, each with $\sim$11 frames, from over 600 raw UAV flight clips through rigorous filtering, excluding content with ambiguous coordinates, extreme scales, or unreliable DA3 reconstructions. The final dataset demonstrates substantial geographical diversity, spanning more than 80 landmarks across 50 cities in approximately 30 countries. Detailed statistics and the complete processing pipeline are provided in the supplementary material.

\section{Experiments}

\subsection{Evaluation Protocol and Implementation Details}
\paragraph{Evaluation Protocol.}
We benchmark representative metric depth estimators on our aerial datasets, including \emph{\Oblique}, \emph{\Decoupled}, and \emph{\Wild}.
Specifically, we evaluate the state-of-the-art methods, including ZoeDepth-NK~\cite{bhat2023zoedepth}, DepthPro~\cite{Bochkovskii2024:arxiv}, UniDepthV1-L~\cite{piccinelli2024unidepth}, UniDepthV2-L~\cite{piccinelli2025unidepthv2}, MoGe2-L~\cite{wang2026moge}, and Metric3Dv2-L~\cite{hu2024metric3d}.
For evaluation, predicted depth is clipped to 0.001--400~m, and invalid pixels are excluded. 
We report AbsRel (the lower the better) and threshold accuracy scores $\delta_{1}$, $\delta_{2}$ (the higher the better).
All metrics are computed per image and averaged over each dataset. We use the publicly available code of the existing methods for testing. 

\paragraph{Model Adaptation Details.}
To validate adaptation on existing methods, we use MoGe2~\cite{wang2026moge} as base model, and we adopt LoRA~\cite{hu2022lora} for parameter-efficient fine-tuning.
The fine-tuned model is denoted as MoGe2-Aerial.
For training data, we use the \emph{\Oblique} training split together with \emph{\Synthetic}. We additionally include ground-domain data (\ie, Hypersim~\cite{roberts2021hypersim}, the MVS-Synth dataset introduced in DeepMVS~\cite{huang2018deepmvs}, and TartanAir~\cite{2020TartanAir}) to better preserve generalization on ground-level scenes.

MoGe2-Aerial is optimized using AdamW, with input images resized to 1K resolution and cropped to dimensions divisible by 14. It is trained with a batch size of 32 for 1,800 steps.
The learning rates for the encoder and decoder are set to $1\times10^{-6}$ and $5\times10^{-5}$, respectively, with polynomial decay.  
Discussion of fine-tuning strategy and data composition are provided in the ablation study.

\begin{table*}[t]
\centering
\caption{Quantitative evaluation of metric depth estimation on aerial datasets.}
\label{tab:metric_depth_comparison}

\resizebox{\textwidth}{!}{
\begin{tabular}{l c c c c c c c c c c c c c c}
\toprule
\multirow{3}{*}{\textbf{Method}} & \multicolumn{6}{c}{\textbf{\Oblique}} & \multicolumn{8}{c}{\textbf{\Decoupled}} \\
\cmidrule(lr){2-7} \cmidrule(lr){8-15}
& \multicolumn{2}{c}{City} & \multicolumn{2}{c}{Natural} & \multicolumn{2}{c}{Rural} & \multicolumn{2}{c}{Building} & \multicolumn{2}{c}{Factory} & \multicolumn{2}{c}{Farm} & \multicolumn{2}{c}{Lawn} \\
\cmidrule(lr){2-3} \cmidrule(lr){4-5} \cmidrule(lr){6-7} \cmidrule(lr){8-9} \cmidrule(lr){10-11} \cmidrule(lr){12-13} \cmidrule(lr){14-15}
& AbsRel & $\delta_{1}$ & AbsRel & $\delta_{1}$ & AbsRel & $\delta_{1}$ & AbsRel & $\delta_{1}$ & AbsRel & $\delta_{1}$ & AbsRel & $\delta_{1}$ & AbsRel & $\delta_{1}$ \\
\midrule
\midrule[-0.5ex]
\rowcolor{gray!15} \multicolumn{15}{c}{\scriptsize \textbf{Metric depth map (w/o GT intrinsics)}} \\
\midrule[-0.5ex]
ZoeDepth~\cite{bhat2023zoedepth} & 97.1 & 0.0 & 93.0 & 0.0 & 92.8 & 0.0 & 95.4 & 0.0 & 96.3 & 0.0 & 96.6 & 0.0 & 95.2 & 0.0 \\

DepthPro~\cite{Bochkovskii2024:arxiv} & 97.8 & 0.0 & 92.2 & 0.0 & 96.9 & 0.0 & 97.8 & 0.0 & 97.6 & 0.0 & 97.3 & 0.0 & 96.7 & 0.0 \\

UniDepthV1~\cite{piccinelli2024unidepth} & 83.9 & 0.0 & 80.3 & 0.0 & 86.0 & 0.0 & 87.4 & 0.0 & 81.1 & 0.0 & 89.5 & 0.0 & 84.6 & 0.0\\
UniDepthV2~\cite{piccinelli2025unidepthv2} & 31.0 & 34.1 & 19.2 & 73.0 & 28.1 & 24.2 & 24.1 & 35.0 & 22.0 & 46.7 & \textbf{17.2} & \textbf{57.9} & 24.4 & 33.8\\

MoGe2~\cite{wang2026moge}  & 48.4 & 5.1 & 45.9 & 23.0 & 56.7 & 0.3 & 29.0 & 26.6 & 26.8 & 34.8 & 71.0 & 0.0 & 32.1 & 19.5 \\

MoGe2-Aerial & \textbf{10.3} & \textbf{89.3} & \textbf{16.8} & \textbf{73.6} & \textbf{14.2} & \textbf{81.2} & \textbf{11.9} & \textbf{87.6} & \textbf{11.4} & \textbf{90.2} & 22.8 & 54.2 & \textbf{9.8} & \textbf{94.0} \\

\midrule[-0.5ex]
\rowcolor{gray!15} \multicolumn{15}{c}{\scriptsize \textbf{Metric depth map (w/ GT intrinsics)}} \\
\midrule[-0.5ex]
UniDepthV1~\cite{piccinelli2024unidepth} & 85.1 & 0.0 & 82.3 & 0.0 & 87.2 & 0.0 & 88.4 & 0.0 & 82.4 & 0.0 & 90.3 & 0.0 & 85.6 & 0.0\\
UniDepthV2~\cite{piccinelli2025unidepthv2} & 29.6 & 34.4 & 17.9 & 76.8 & 18.4 & 57.6 & 21.0 & 43.5 & 18.2 & 61.0 & \textbf{10.9} & \textbf{91.0} & 18.4 & 57.4\\
DepthAnything 3~\cite{lin2025depth3} & 96.6 & 0.0 & 84.9 & 0.0 & 91.3 & 0.0 & 87.0 & 0.0 & 93.3 & 0.0 & 93.1 & 0.0 & 93.9 & 0.0\\

Metric3DV2~\cite{hu2024metric3d} & 75.4 & 0.0 & 68.5 & 0.1 & 73.2 & 0.1 & 68.5 & 0.0 & 63.9 & 0.0 & 81.1 & 0.0 & 73.3 & 0.0 \\

MoGe2~\cite{wang2026moge}  & 36.2 & 18.1 & 43.4 & 25.1 & 42.3 & 8.0 & 19.1 & 59.7 & 12.3 & 83.4 & 61.9 & 0.8 & 18.5 & 61.0 \\

MoGe2-Aerial  & \textbf{8.8} & \textbf{93.8} & \textbf{14.3} & \textbf{78.5} & \textbf{11.8} & \textbf{85.3} & \textbf{10.9} & \textbf{90.3}  & \textbf{13.9} & \textbf{85.7} & 17.8 & 71.2 & \textbf{11.5}  & \textbf{90.7} \\

\bottomrule
\end{tabular}
}

    \centering
    \includegraphics[width=\textwidth]{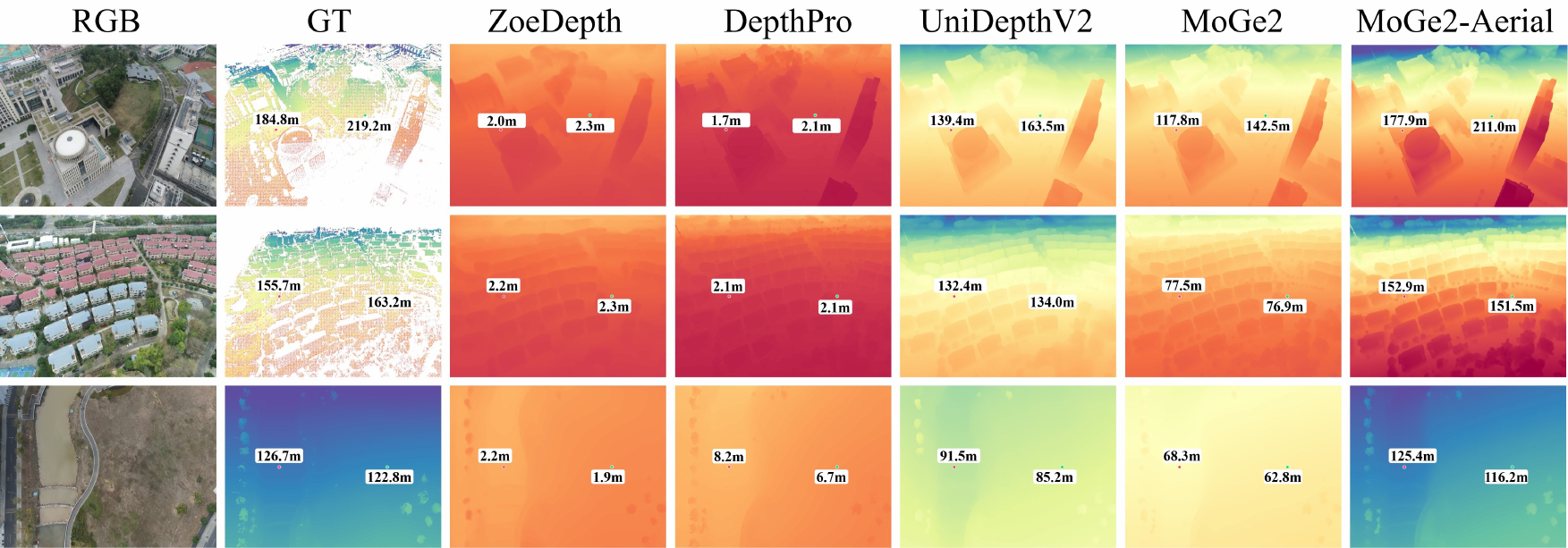}
    \captionof{figure}{Qualitative comparison with representative methods on aerial scenes.}
    \label{fig:qual_sota}
\end{table*}

\subsection{Benchmarking Metric Depth Estimation Methods}
\paragraph{Evaluation on \Oblique and \Decoupled.}\Tref{tab:metric_depth_comparison} presents the metric depth prediction results with and without ground-truth camera intrinsics.
Across both protocols, zero-shot transfer from ground-domain pretraining to aerial imagery is weak, with several baselines showing very high AbsRel and near-zero $\delta_{1}$, indicating a severe domain gap.
In zero-shot evaluation, UniDepthV2\cite{piccinelli2025unidepthv2} is the strongest baseline, but still shows clear gaps and instability across scenes. 

Adapting existing method with our \emph{\DatasetName} consistently improves performance. 
\Tref{tab:metric_depth_comparison} shows that MoGe2-Aerial significantly outperforms zero-shot MoGe2\cite{wang2026moge} on all aerial datasets. For example, on \emph{\Oblique-City} without ground-truth intrinsic, $\delta_{1}$ increases from 5.1\% to 89.3\%.

\begin{figure*}[t]
    \centering
    \includegraphics[width=\textwidth]{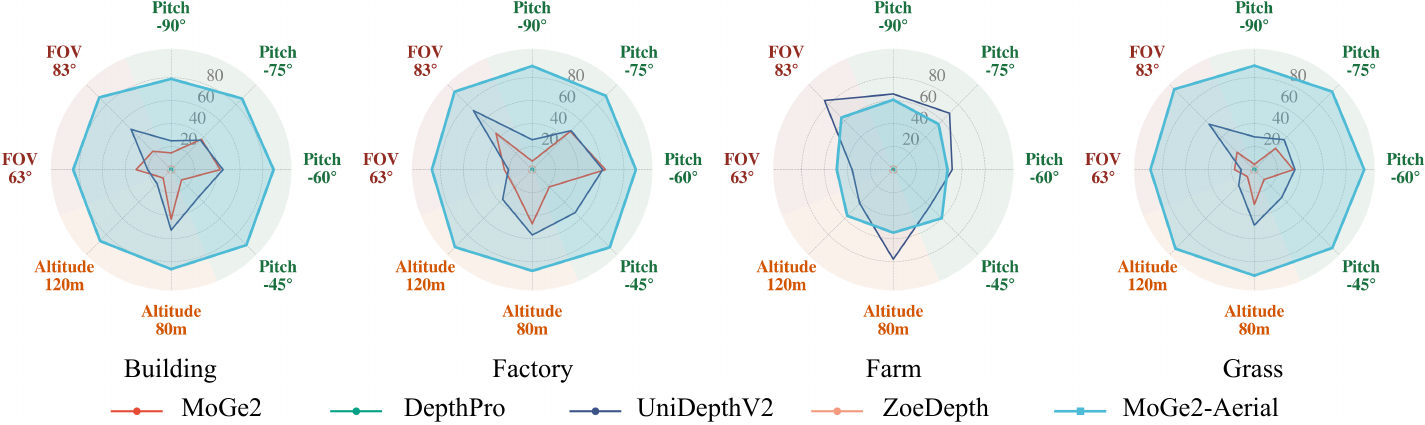}
    \caption{Radar summary of parameter robustness on \Decoupled. Larger radius indicates better performance under each decoupled setting.}
    \label{fig:param_radar}
\end{figure*}

The qualitative results in \fref{fig:qual_sota} support these trends. Compared with ZoeDepth~\cite{bhat2023zoedepth}, DepthPro~\cite{Bochkovskii2024:arxiv}, UniDepthV2~\cite{piccinelli2025unidepthv2},  and MoGe2~\cite{wang2026moge} our adapted model produces cleaner boundaries, fewer local depth inversions, and more coherent global scale, especially in distant roads, and vegetation regions.

\paragraph{Hard scene analysis.}
The \emph{Farm} subset in \emph{\Decoupled} remains the most challenging setting.
MoGe2-Aerial exhibits lower accuracy on this subset compared to others (\eg, $\delta_{1}=54.2\%$ in \Tref{tab:metric_depth_comparison}). This performance bottleneck is likely attributable to the prevalence of repetitive vegetation textures and a lack of strong geometric anchors. Interestingly, UniDepthV2~\cite{piccinelli2025unidepthv2} achieves the highest performance on the \emph{Farm} category. 
We hypothesize that this occurs because UniDepthV2's pre-training corpus encompasses a data distribution resembling agricultural environments.

\paragraph{Robustness Under Decoupled Camera Factors}
We systematically evaluate the robustness of our method against variations in camera pitch, altitude, and FOV. As illustrated by the radar summaries in Fig. 5, MoGe2-Aerial exhibits an expansive and highly isotropic performance footprint across diverse scene types (\emph{Building}, \emph{Factory}, \emph{Farm}, and \emph{Lawn}). This demonstrates remarkable resilience to aerial-specific camera factors, whereas prior foundation models yield irregular and contracted footprints, indicating severe sensitivity to these parameters.

Figure 6 further details the impact of FOV and altitude shifts. Zero-shot foundation models (e.g., MoGe2\cite{wang2026moge}, UniDepthV2\cite{piccinelli2025unidepthv2}) experience a precipitous drop in metric accuracy when the altitude increases from 80m to 120m. Although a wider FOV ($83^\circ$ compared to $63^\circ$) partially mitigates this degradation by capturing broader spatial context, the overall baseline performance remains inadequate. In contrast, MoGe2-Aerial effectively resolves the scale ambiguity prevalent in high-altitude captures, maintaining consistently high accuracy (over $80\%$) across all FOV and altitude configurations.

\begin{figure*}[t]
    \centering
    \begin{minipage}[c]{0.48\textwidth} %
        \caption{Robustness analysis of depth estimation models (MoGe2, UniDepthV2, and MoGe2-Aerial) across different Field of Views (FOV) and altitudes. Results represent the average $\delta_{1}$ on the \Decoupled dataset.}
        \label{fig:depth_robustness}
    \end{minipage}
    \hfill
    \begin{minipage}[c]{0.48\textwidth}
        \centering
        \includegraphics[width=0.9\textwidth]{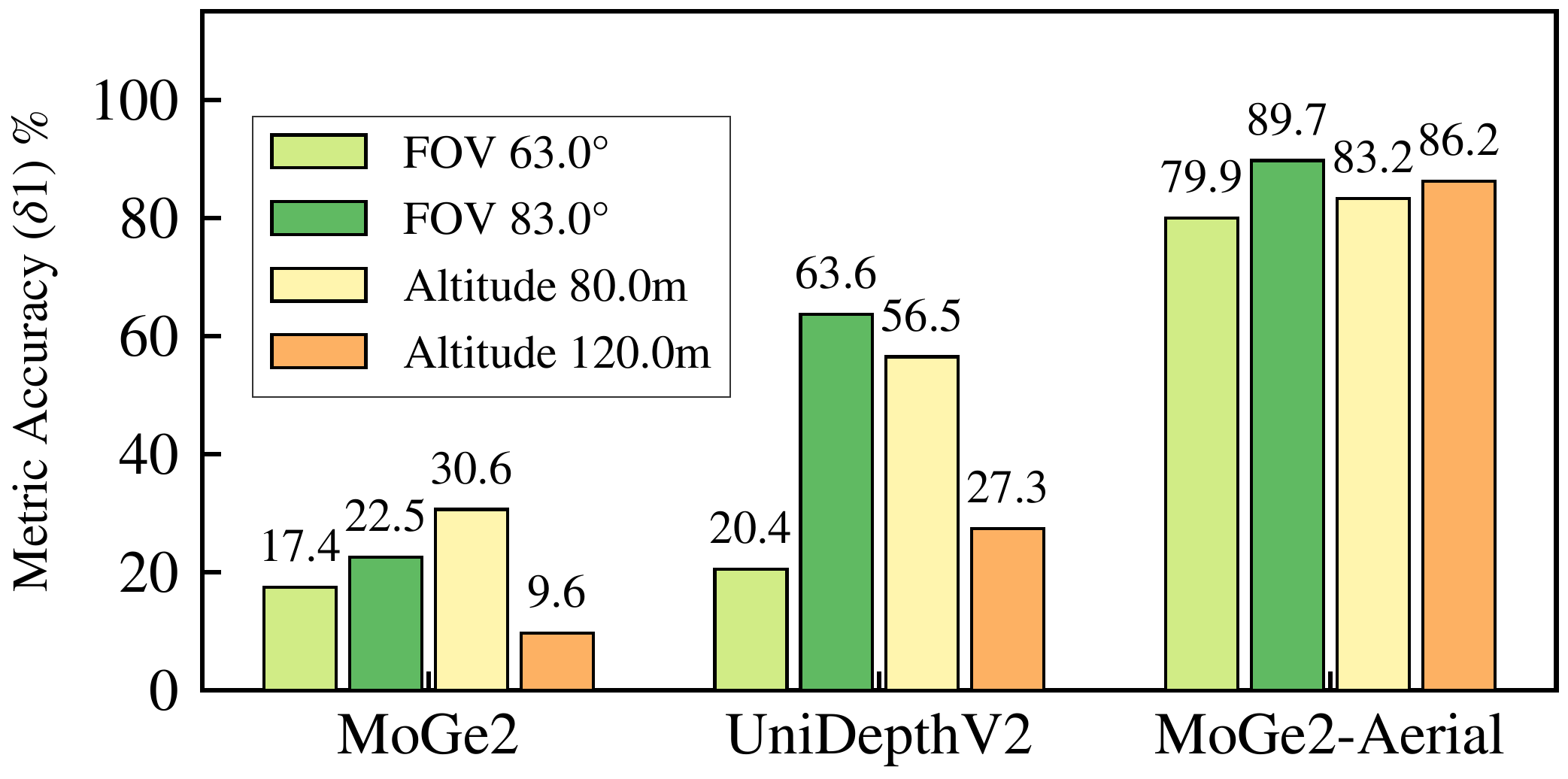}
    \end{minipage}

    \centering
    \includegraphics[width=0.95\textwidth]{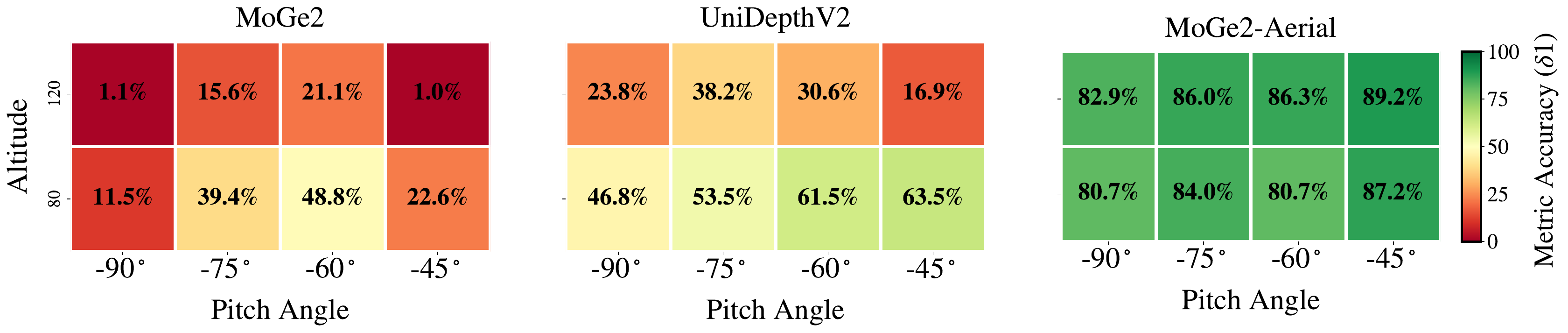} 
    \caption{Heatmap visualization of metric accuracy across altitudes and pitch angles.}
    \label{fig:heatmap_analysis}
\end{figure*}

Furthermore, the heatmaps in \fref{fig:heatmap_analysis} highlight the joint influence of altitude and camera pitch. Baseline models exhibit severe, non-monotonic degradation regarding pitch, suffering catastrophic failures at both strictly nadir ($-90^\circ$) and oblique views up to ($-45^\circ$) viewing angles. This vulnerability stems directly from the domain gap between ground-centric training data and aerial perspectives. Conversely, our fine-tuned MoGe2-Aerial demonstrates exceptional stability, retaining uniformly high accuracy across all evaluated pitch angles and altitudes without succumbing to the domain shifts that hinder zero-shot approaches.

\paragraph{Generalization to in-the-wild videos.}
We further evaluate existing methods on \emph{\Wild}, which contains unconstrained internet aerial videos with large appearance variation and wide depth ranges. This setting is particularly challenging, and existing zero-shot models still show high AbsRel and low threshold accuracy (see \Tref{tab:range_comparison}).
After fine-tuning on \emph{\DatasetName}, performance improves markedly. Compared with zero-shot MoGe2\cite{wang2026moge}, our adapted model reduces AbsRel from 55.26 to 21.34 in the 0--400\,m range, while improving $\delta_{1}$ from 9.5 to 53.7, respectively. These results demonstrate that \emph{\DatasetName} can enhance both robustness and generalization in monocular aerial depth estimation.

\subsection{Evaluation on Ground-Domain Datasets}
\begin{wraptable}{r}{0.45\textwidth} 
    \centering
    \caption{Result on \Wild.
    }
    \label{tab:range_comparison}

\resizebox{\linewidth}{!}{
    \begin{tabular}{l ccc ccc}
    \toprule
    \multirow{2}{*}{\textbf{Method}} & \multicolumn{3}{c}{\textbf{Range: 0--400m}} & \multicolumn{3}{c}{\textbf{Range: 0--800m}} \\
    \cmidrule(lr){2-4} \cmidrule(lr){5-7}
    & AbsRel  & $\delta_{1}$  & $\delta_{2}$  & AbsRel  & $\delta_{1}$  & $\delta_{2}$  \\
    \midrule
    ZoeDepth\cite{bhat2023zoedepth} & 93.56 & 0.27 & 1.03 & 94.10 & 0.26 & 0.99\\
    DepthPro\cite{Bochkovskii2024:arxiv} & 98.06 & 0.00 & 0.02 & 98.18 & 0.00 & 0.03\\
    UniDepthV2\cite{piccinelli2025unidepthv2} & 52.88 & 27.65 & 61.34 & 55.97 & 22.92 & 55.02 \\
    Metric3Dv2\cite{hu2024metric3d} & 82.38 & 4.90 & 13.90 & 85.31 & 4.82 & 11.18 \\
    
    MoGe2\cite{wang2026moge} & 55.26 & 9.50 & 21.40 & 55.56 & 9.70 & 21.60 \\
    MoGe2-Aerial & \textbf{21.34} & \textbf{53.7} & \textbf{85.1} & \textbf{22.28} & \textbf{52.3} & \textbf{83.9} \\
    \bottomrule
    \end{tabular}
}

\end{wraptable}

To verify that aerial adaptation does not overfit only to \emph{\Oblique} and \emph{\Decoupled}, we additionally report cross-domain performance on seven ground benchmarks (NYUv2\cite{2012nyu}, KITTI\cite{geiger2013vision}, ETH3D\cite{schops2017ETH3D}, iBims\cite{koch2018IBIMS}, DDAD\cite{guizilini2020DDAD}, DIODE\cite{vasiljevic2019diode}, HAMMER\cite{jung2023HAMMER}) to measure transfer and forgetting after aerial adaptation.

\Tref{tab:ground_depth_comparison} shows that MoGe2-Aerial remains competitive on ground benchmarks and improves multiple datasets over zero-shot MoGe2\cite{wang2026moge} (e.g., KITTI, and iBims, DDAD), supporting that the adapted model preserves generalization beyond aerial domains.

\begin{table*}[t]
\centering
\caption{Quantitative evaluation of metric depth estimation on ground datasets.}
\label{tab:ground_depth_comparison}
\resizebox{\textwidth}{!}{
\begin{tabular}{l c c c c c c c c c c c c c c}
\toprule
\multirow{2}{*}{\textbf{Method}} & \multicolumn{2}{c}{\textbf{NYUv2}} & \multicolumn{2}{c}{\textbf{KITTI}} & \multicolumn{2}{c}{\textbf{ETH3D}} & \multicolumn{2}{c}{\textbf{iBims}} & \multicolumn{2}{c}{\textbf{DDAD}} & \multicolumn{2}{c}{\textbf{DIODE}} & \multicolumn{2}{c}{\textbf{HAMMER}} \\
\cmidrule(lr){2-3} \cmidrule(lr){4-5} \cmidrule(lr){6-7} \cmidrule(lr){8-9} \cmidrule(lr){10-11} \cmidrule(lr){12-13} \cmidrule(lr){14-15}
& AbsRel & $\delta_1$ & AbsRel & $\delta_1$ & AbsRel & $\delta_1$ & AbsRel & $\delta_1$ & AbsRel & $\delta_1$ & AbsRel & $\delta_1$ & AbsRel & $\delta_1$ \\
\midrule
\midrule[-0.5ex]
\rowcolor{gray!15} \multicolumn{15}{c}{\scriptsize \textbf{Metric depth map (w/o GT intrinsics)}} \\
\midrule[-0.5ex]
ZoeDepth\cite{bhat2023zoedepth} & 11.0 & 91.9 & 17.0 & 85.4 & 57.1 & 33.7 & 17.4 & 67.2 & 38.9 & 38.6 & 39.3 & 29.3 & 94.3 & 3.23 \\

DepthPro\cite{Bochkovskii2024:arxiv} & 10.7 & 91.9 & 23.5 & 38.3 & 38.5 & 32.8 & 15.9 & 81.5 & 33.4 & 35.3 & 31.9 & 37.7 & 39.1 & 63.0 \\
UniDepthV2\cite{piccinelli2025unidepthv2} & 10.2 & 93.4 & \textbf{9.1} & \textbf{91.4} & 20.7 & 69.5 & \textbf{9.0} & \textbf{93.1} & 18.4 & \textbf{77.5} & 42.9 & 51.7 & 38.1 & 46.8 \\

MoGe2\cite{wang2026moge} & \textbf{6.9} & \textbf{96.7} & 17.6 & 64.7 & \textbf{10.0} & \textbf{88.8} & 14.6 & 80.5 & \textbf{15.6} & 73.7 & 16.7 & 67.9 & \textbf{26.0} & 65.3 \\
MoGe2-Aerial & 7.1 & 96.2 & 16.3 & 68.7 & 13.1 & 86.7 & 14.7 & 85.4 & 16.2 & 74.3 & \textbf{15.1} & \textbf{77.0} & 31.7 & \textbf{68.9} \\
\midrule[-0.5ex]
\rowcolor{gray!15} \multicolumn{15}{c}{\scriptsize \textbf{Metric depth map (w/ GT intrinsics)}} \\
\midrule[-0.5ex]

UniDepthV2\cite{piccinelli2025unidepthv2} & 7.5 & 96.2 & 5.6 & 97.5 & 15.0 & 85.2 & 7.8 & 94.7 & 14.1 & 89.2 & 40.9 & 67.0 & 37.7 & 47.1 \\
DepthAnything 3\cite{lin2025depth3} & 7.1 & 97.0 & 7.3 & 96.0 & 12.5 & 85.5 & \textbf{7.5} & \textbf{96.7} & 12.4 & 87.3 & 15.2 & 76.6 & \textbf{27.4} & 66.0 \\
Metric3Dv2\cite{hu2024metric3d} & 7.2 & 96.5 & \textbf{5.3} & \textbf{98.0} & 11.8 & 88.8 & 9.96 & 94.1 & \textbf{9.21} & \textbf{93.7} & 49.1 & 1.98 & 35.7 & 44.3 \\

MoGe2\cite{wang2026moge} & \textbf{6.0} & \textbf{97.2} & 8.6 & 93.8 & \textbf{9.3} & \textbf{93.7} & 10.4 & 92.0 & 13.0 & 85.6 & \textbf{14.9} & 80.1 & 29.8 & \textbf{72.5} \\
MoGe2-Aerial & 6.0 & 97.1 & 8.1 & 94.3 & 12.9 & 85.5 & 10.9 & 91.8 & 12.9 & 86.7 & 15.1 & \textbf{81.1} & 35.8 & 72.0 \\
\bottomrule
\end{tabular}
}

\end{table*}

\subsection{Ablation Study}
\begin{figure*}[t]
    \centering
    \includegraphics[width=\textwidth]{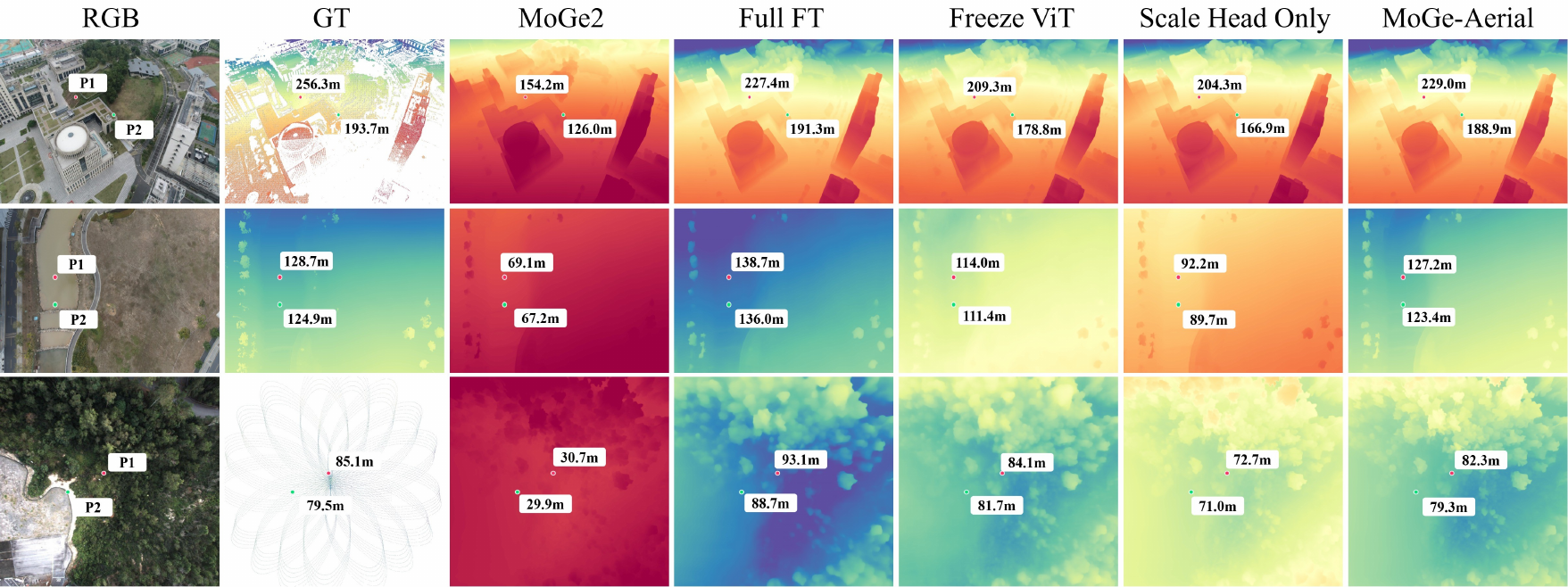} 
    \caption{Qualitative comparison of fine-tuning strategies.}
    \label{fig:finetune_strategy}
\end{figure*}

\paragraph{Fine-tuning Strategy Analysis.} 
To determine the most effective adaptation approach, we evaluate several fine-tuning strategies: full fine-tuning, partial fine-tuning (freezing the ViT backbone or updating only the scale head), and Low-Rank Adaptation (LoRA) with varying ranks. For LoRA, we inject adapters into all linear layers (\texttt{qkv}, \texttt{proj}, \texttt{fc1}, and \texttt{fc2}) with $r \in \{64, 96, 128\}$.

As shown in Table~\ref{tab:ablation_finetuning_full}, while full fine-tuning achieves the highest scores on aerial datasets, it significantly degrades generalization to ground-level scenes. Partial fine-tuning, either freezing the backbone or isolating updates to the scale head—yields suboptimal performance. In contrast, LoRA provides a superior trade-off; specifically, $r=96$ offers the best balance across diverse domains. Increasing the rank to $r=128$ results in a slight performance drop, likely due to over-parameterization or increased sensitivity to distribution noise.

\begin{table}[t] %
    \centering
    \begin{minipage}{0.40\textwidth} %
        \centering
        \caption{Fine-tuning strategies.}
        \label{tab:ablation_finetuning_full}
        \resizebox{\linewidth}{!}{
    \begin{tabular}{l c c c c c c}
    \toprule
    \multirow{2}{*}{\textbf{Strategy}} & \multicolumn{2}{c}{\textbf{Oblique}} & \multicolumn{2}{c}{\textbf{Decoupled}} & \multicolumn{2}{c}{\textbf{Ground}} \\
    \cmidrule(lr){2-3} \cmidrule(lr){4-5} \cmidrule(lr){6-7}
    & AbsRel & $\delta_{1}$ & AbsRel & $\delta_{1}$ & AbsRel & $\delta_{1}$ \\
    \midrule
    Zero-shot & 49.50 & 7.90 & 38.38 & 20.00 & \textbf{15.34} & 76.78 \\
    Full FT & \textbf{6.68} & \textbf{95.60} & 13.22 & \textbf{86.00} & 35.55 & 56.70 \\
    Freeze ViT & 10.81 & 87.70 & 17.19 & 82.50 & 37.72 & 35.80 \\
    Scale Head & 22.85 & 53.00 & 15.22 & 71.90 & 43.72 & 36.00 \\
    \midrule
    LoRA ($r=64$) & 12.66 & 82.90 & 13.03 & 84.30 & 16.02 & \textbf{80.70} \\
    LoRA ($r=96$) & 12.45 & 84.40 & 13.19 & 83.90 & 16.30 & 79.59 \\
    LoRA ($r=128$)& 12.94 & 82.40 & \textbf{13.00} & 84.20 & 18.80 & 69.40 \\
    \bottomrule
    \end{tabular}
}

    \end{minipage}
    \hfill %
    \begin{minipage}{0.58\textwidth} %
        \centering
        \caption{Analysis of training dataset composition.}
        \label{tab:data_ablation_grouped_baselines}

\resizebox{\textwidth}{!}{
\begin{tabular}{l c c c c c c c c}
\toprule
\multirow{3}{*}{\textbf{Training Data}} & \multicolumn{4}{c}{\textbf{Aerial}} & \multicolumn{4}{c}{\textbf{Ground}} \\
\cmidrule(lr){2-5} \cmidrule(lr){6-9}

& \multicolumn{2}{c}{\textbf{Oblique (Mean)}} & \multicolumn{2}{c}{\textbf{Decoupled (Mean)}} & \multicolumn{2}{c}{\textbf{Ground-domain (Mean)}} & \multicolumn{2}{c}{\textbf{ETH3D}} \\
\cmidrule(lr){2-3} \cmidrule(lr){4-5} \cmidrule(lr){6-7} \cmidrule(lr){8-9}

& AbsRel & $\delta_{1}$ & AbsRel & $\delta_{1}$ & AbsRel & $\delta_{1}$ & AbsRel & $\delta_{1}$ \\
\midrule %

MoGe2 & 49.50 & 7.90 & 38.38 & 20.00 & \textbf{15.34} & 76.78 & \textcolor{blue}{\textbf{9.98}} & \textcolor{blue}{\textbf{88.79}} \\
\midrule
Oblique & \textbf{11.42} & \textbf{87.4} & 16.65 & 77.45 & 20.51 & 77.64 & \textcolor{blue}{26.37} & \textcolor{blue}{61.50} \\

Oblique+Synth& 12.69 & 82.30 & \textbf{13.05} & \textbf{84.20} &  17.68 & \textbf{79.65} & \textcolor{blue}{19.99} & \textcolor{blue}{71.54} \\ 

Oblique+Synth+Ground & 12.45 & 84.40 & 13.19 & 83.90 & 16.30 & 79.59 & \textcolor{blue}{13.10} & \textcolor{blue}{86.67} \\

\bottomrule
\end{tabular}
}

    \end{minipage}
\end{table}

\paragraph{Effect of Training Data Composition.}
In the single-stage mixed-domain training phase, we balance the influence of various data sources through weighted sampling. The sampling proportions for \emph{\Oblique}, \emph{\Synthetic}, and Ground-domain datasets are maintained at $80\%$, $15\%$, and $5\%$, respectively.

Table~\ref{tab:data_ablation_grouped_baselines} shows the effect of multi-domain data composition. Training with only aerial-domain data gives strong gains in \emph{\Oblique} but leads to severe catastrophic forgetting on ground-domain tasks, as evidenced by a substantial performance drop on the ground dataset. While adding \emph{\Synthetic} improves results on \emph{\Decoupled}, it is insufficient to fully recover the robustness on ground-domain performance, with $\delta_1$ results on ETH3D degrade to $71.54$. 
Incorporating a small fraction of ground data effectively mitigates this forgetting issue, limiting the $\delta_{1}$ degradation on ETH3D to merely $5.1$ compared to the initial baseline, ultimately yielding the best overall compromise. Detailed results can be found in the supplementary materials.

\section{Conclusion}
\label{sec:Conclusion}

We present \DatasetName, a large-scale and diverse benchmark for monocular metric depth estimation in aerial UAV imagery.
Our dataset combines four complementary subsets, including large-scale real and synthetic image-depth pairs, a controlled acquisition subset for factorized analysis, and in-the-wild aerial imagery for cross-domain evaluation.
Built on this benchmark, we conduct systematic studies of how viewpoint, altitude, and camera parameters affect aerial metric depth prediction, and establish a comprehensive evaluation protocol for this setting.
By fine-tuning representative state-of-the-art models on \DatasetName, we achieve substantial gains and set new state-of-the-art performance across diverse UAV deployment scenarios.

\paragraph{Future Work}
An important next step is extending the current monocular benchmark to multi-view metric depth estimation, enabling stronger geometric consistency and broader applicability to UAV mapping and reconstruction pipelines.
In addition, although our dataset improves aerial diversity, it still has limited coverage of extreme weather conditions, and future data collection under adverse weather is needed to improve robustness in real-world deployments.

\bibliographystyle{splncs04}
\bibliography{ref}
\clearpage

\title{Supplementary Material for \\ ``AerialMetric: Benchmarking and Adapting UAV Monocular Metric Depth Estimation in the Real World''}

\author{
Zhongqiang Song\inst{1} \and
Guanying Chen\inst{1}\thanks{Corresponding author.} \and
Yuqi Zhang\inst{2,3} \and
Yin Zou\inst{1} \and
Chuanyu Fu\inst{1} \and
Zhiyuan Yuan\inst{1} \and
Chuan Huang\inst{4,2} \and
Shuguang Cui\inst{3,2} \and
Xiaochun Cao\inst{1}
}

\authorrunning{Z. Song et al.}

\institute{%
$^{1}$Sun Yat-sen University, Shenzhen Campus \quad
$^{2}$FNii--Shenzhen \quad \\
$^{3}$SSE, CUHKSZ \quad
$^{4}$SIAS, UESTC}

\titlerunning{AerialMetric: UAV Metric Depth Estimation}
\maketitle
\appendix

\section{More Details for the Dataset}

Our proposed \emph{AerialMetric} dataset contains over 68,000 image--depth
pairs from 129 real-world scenes and 7 synthetic environments, substantially
larger than existing aerial depth datasets such as
UseGeo~\cite{nex2024usegeo} (829 pairs),
WildUAV~\cite{florea2021wilduav} (1,500 pairs), and
OccuFly~\cite{gross2026occufly} (20,160 pairs). The benchmark is organized
into four specialized sub-datasets: \emph{\Oblique}, \emph{\Decoupled},
\emph{\Synthetic}, and \emph{\Wild}. Detailed statistics and scene attributes
are summarized in \Tref{tab:dataset_stats}.

\begin{figure}[t] \centering \includegraphics[width=\linewidth]{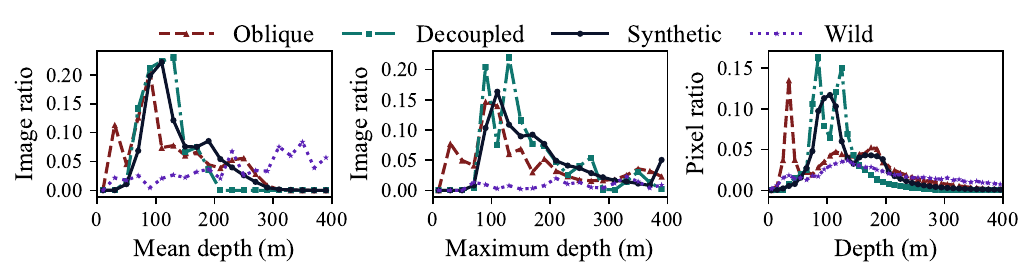}\\  \caption{Depth statistics of the proposed datasets.} \label{fig:depth_statistics}  \end{figure}

Supplementary statistics further provide a finer-grained characterization of
the acquisition conditions, including altitude, scene category, FOV, and
pitch angle. Fig.~\ref{fig:depth_statistics} additionally visualizes the
distributions of per-image mean depth, maximum depth, and pixel-wise depth,
highlighting the broad geometric and depth coverage of \emph{AerialMetric}.

\begin{table}[!htbp] %
\centering
\caption{Detailed statistics and scene attributes for each AerialMetric sub-dataset.}
\label{tab:dataset_stats}

\resizebox{\textwidth}{!}{%
\setlength{\tabcolsep}{8pt} %
\begin{tabular}{@{}llrcccc@{}} %
\toprule
\textbf{Source} & \textbf{Scene} & \textbf{Images} & \textbf{Area (m²)} & \textbf{Resolution} & \textbf{Altitude (m)} & \textbf{Category} \\ 
\midrule

\multicolumn{7}{c}{\textbf{AerialMetric-Oblique}} \\
\midrule %
\multirow{2}{*}{UrbanBIS~\cite{UrbanBIS}} 
& Longhua    & 9,295 &$2.4 \times 10^6$ & $8192 \times 5460$ & 170 & City \\
& Yingrenshi &   854 &$8.0 \times 10^4$ & $5472 \times 3468$ &  35 & City \\ 
\addlinespace

\multirow{6}{*}{GauU-SceneV2~\cite{xiong2024gauuscenev2assessingreliability}} 
& LFLS       & 1,099 &$1.4 \times 10^6$ & $5472 \times 3468$ & 150 & City \\
& Upper      &   715 &$9.2 \times 10^5$ & $5472 \times 3468$ & 120 & City \\
& SZIIT      & 1,215 &$1.5 \times 10^6$ & $5472 \times 3468$ & 136 & City \\
& SZTU       & 1,500 & -   & $5472 \times 3468$ & 180 & City \\
& SMBU       &   561 &$9.0 \times 10^5$ & $5472 \times 3468$ & 150 & City \\
& HAV        &   393 &$8.1 \times 10^5$ & $5472 \times 3468$ & 120 & City \\ 
\addlinespace

\multirow{4}{*}{UAVScenes~\cite{wang2025uavscenes}} 
& Town       & 3,892 &$1.1 \times 10^5$ & $2448 \times 2048$ & 80 & Rural \\
& Valley     & 3,560 &$1.1 \times 10^5$ & $2448 \times 2048$ & 80 & Natural \\
& Airport    & 3,044 &$1.1 \times 10^5$ & $2448 \times 2048$ & 80 & Rural\\
& Island     & 4,025 &$1.0 \times 10^5$ & $2448 \times 2048$ & 80 & Natural \\
\addlinespace

\multirow{4}{*}{ESRI~\cite{esri}} 
& BC-Scene   &   610 &$4.6 \times 10^5$ & $8192 \times 5460$ &  80 & Rural \\
& R-PHD      &   307 &$4.1 \times 10^5$ & $9504 \times 6336$ & 100 & Rural \\
& L-Scene    &   348 &$8.8 \times 10^4$ & $8192 \times 5460$ & 100 & Rural \\
& T-Scene    & 1,020 &$1.1 \times 10^5$ & $9504 \times 6336$ &  30 & Rural \\
\addlinespace

\multirow{7}{*}{OpenDroneMap~\cite{opendronemap}} 
& Bellus     &   115 &$2.3 \times 10^5$ & $4048 \times 3048$ & 140 & Natural \\
& Lewis      &   134 &$6.9 \times 10^5$ & $4000 \times 3000$ &  60 & Natural \\
& ODM1       &   271 &$3.1 \times 10^5$ & $5472 \times 3468$ & 100 & Rural \\
& ODM2       &   297 &$3.1 \times 10^5$ & $5472 \times 3468$ & 100 & Rural \\
& ODM3       &   493 &$5.9 \times 10^5$ & $4000 \times 3000$ & 100 & Natural \\
& Park       & 6,666 &$6.4 \times 10^5$ & $4000 \times 3000$ & 35, 60 & Natural \\
& sceneca    &   166 &$1.4 \times 10^5$ & $3600 \times 2700$ &  90 & Natural \\
\addlinespace

\multirow{2}{*}{UrbanScene3D~\cite{UrbanScene3D}} 
& PolyTech   & 2,368 &$5.9 \times 10^5$ & $6000 \times 4000$ & 120 & City \\
& Artsci     & 3,307 &$8.1 \times 10^5$ & $6000 \times 4000$ & 115 & City \\

\midrule
\multicolumn{7}{c}{\textbf{AerialMetric-Decoupled}} \\
\midrule
\multirow{4}{*}{Self-Collected}            
& Building &   880 & $2.6 \times 10^4$ & \multirow{4}{*}{\begin{tabular}[c]{@{}c@{}}$5472 \times 3648$\\$8192 \times 5460$\end{tabular}} & \multirow{4}{*}{80, 120} & \multirow{4}{*}{-} \\
& Factory  &   948 & $10.0 \times 10^4$& & & \\
& Lawn     & 1,851 & $7.2 \times 10^4$ & & & \\
& Farm     &   946 & $9.0 \times 10^4$ & & & \\

\midrule
\multicolumn{7}{c}{\textbf{AerialMetric-Synthetic}} \\
\midrule
\multirow{2}{*}{UE}            
& Sydney   &  3,816 & $4.0 \times 10^6$ & $3840 \times 2160$ & 80--280 & - \\
& Rural    & 11,907 & $9.0 \times 10^4$ & $3840 \times 2160$ & 80--280 & - \\
\addlinespace

\multirow{5}{*}{GoogleEarthStudio}            
& City     & 400 & $3.5 \times 10^6$ & $3840 \times 2160$ & 80--280 & - \\
& Park     & 400 & $5.9 \times 10^6$ & $3840 \times 2160$ & 80--280 & - \\
& Landmark & 400 & $1.8 \times 10^6$ & $3840 \times 2160$ & 80--280 & - \\
& Factory  & 400 & $5.9 \times 10^6$ & $3840 \times 2160$ & 80--280 & - \\
& Campus   & 400 & $6.1 \times 10^6$ & $3840 \times 2160$ & 80--280 & - \\

\midrule
\multicolumn{7}{c}{\textbf{AerialMetric-Wild}} \\
\midrule
Internet & Wild & 1,100 & - & $1920 \times 1080$ & - & - \\
\bottomrule
\end{tabular}%
}

\end{table}

\subsection{\Oblique: UAV Photogrammetry Datasets}
\begin{figure*}[thbp]
    \centering
    \includegraphics[width=\textwidth]{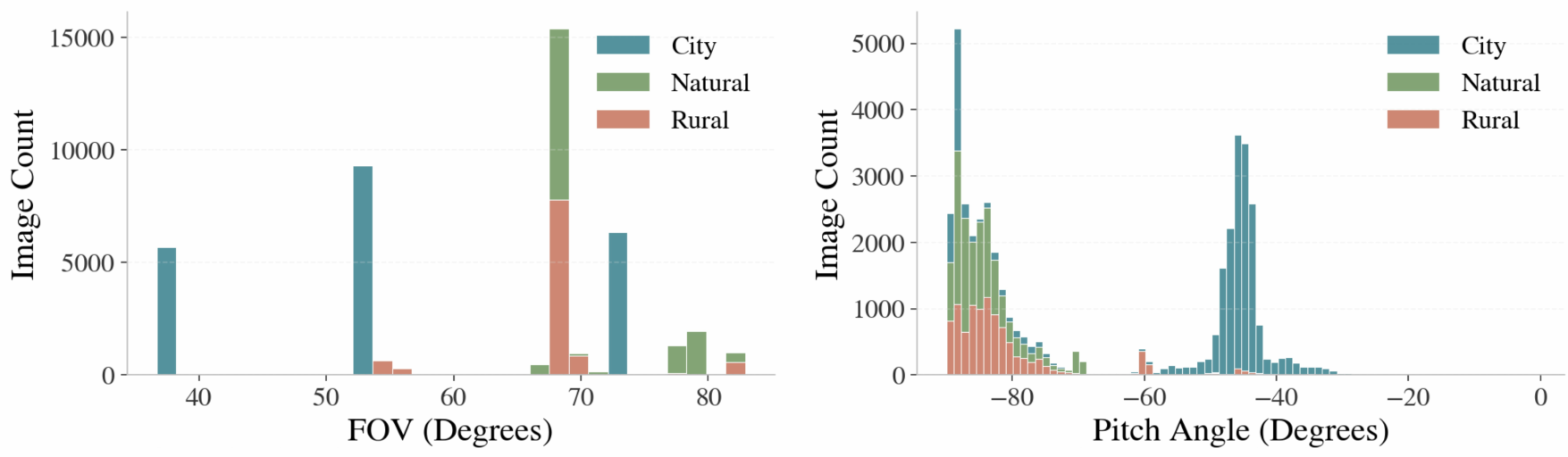} 
    \caption{Statistics of the \Oblique dataset. The left panel shows the image count across various discrete Field of View (FOV) angles. The right panel illustrates the distribution of camera pitch angles.}
    \label{fig:oblique}

    \centering
    \includegraphics[width=1.0\textwidth]{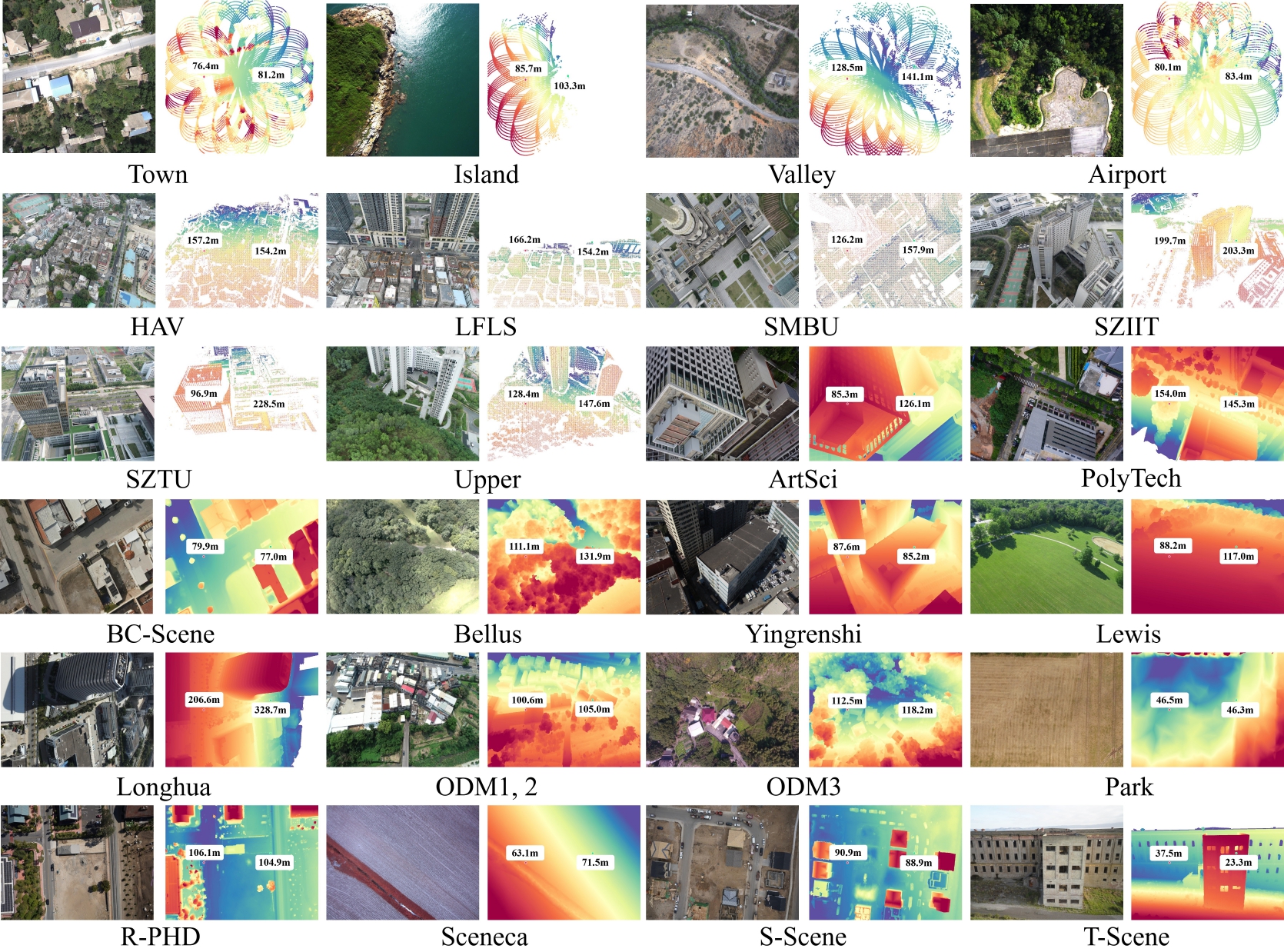} 
    \caption{Visual overview of the \Oblique dataset.}
    \label{fig:all_scenes_vis}
\end{figure*}

\paragraph{Dataset Statistics.}
To provide a granular view of the dataset composition, \cref{fig:oblique} shows the distribution of FOV and pitch angle of different categories. As illustrated, the dataset exhibits extensive diversity by aggregating data from multiple FOVs, a wide range of pitch angles, and various challenging scene types. Furthermore, \cref{fig:all_scenes_vis} visualizes representative RGB inputs and their GT metric depth in these scenarios, ranging from dense urban CBDs to rural landscapes, factories, and synthetic environments. This rich combination of diverse camera configurations and environmental contexts ensures our dataset comprehensively reflects the complexity of real-world aerial scenarios. 

\paragraph{Ground Truth Fidelity and Quality Control.} 
To characterize the precision of our data acquisition pipeline, we provide a quantitative analysis of the \emph{PolyTech} scene as a representative benchmark. 
This large-scale environment was reconstructed from a dense collection of 2,368 high-resolution frames ($6000 \times 4000$ pixels). 
The reconstruction process involved an extensive 640-minute optimization via the DJI Terra\cite{DJITERRA} 3D engine to ensure a high-fidelity, geometrically consistent mesh.

In terms of spatial resolution, the pipeline achieved an average Ground Sampling Distance (GSD) of 1.63~cm, capturing the fine-grained structural features essential for high-precision metric depth supervision. 
Furthermore, the multi-view triangulation attained sub-pixel accuracy, maintaining a mean reprojection error (MRE) of 0.993 pixels. 
These technical specifications exemplify the rigorous standards applied across our dataset, ensuring that the generated ground truth serves as a high-fidelity reference for metric depth evaluation.

\begin{figure}[tbp]
    \centering
    \includegraphics[width=1.0\textwidth]{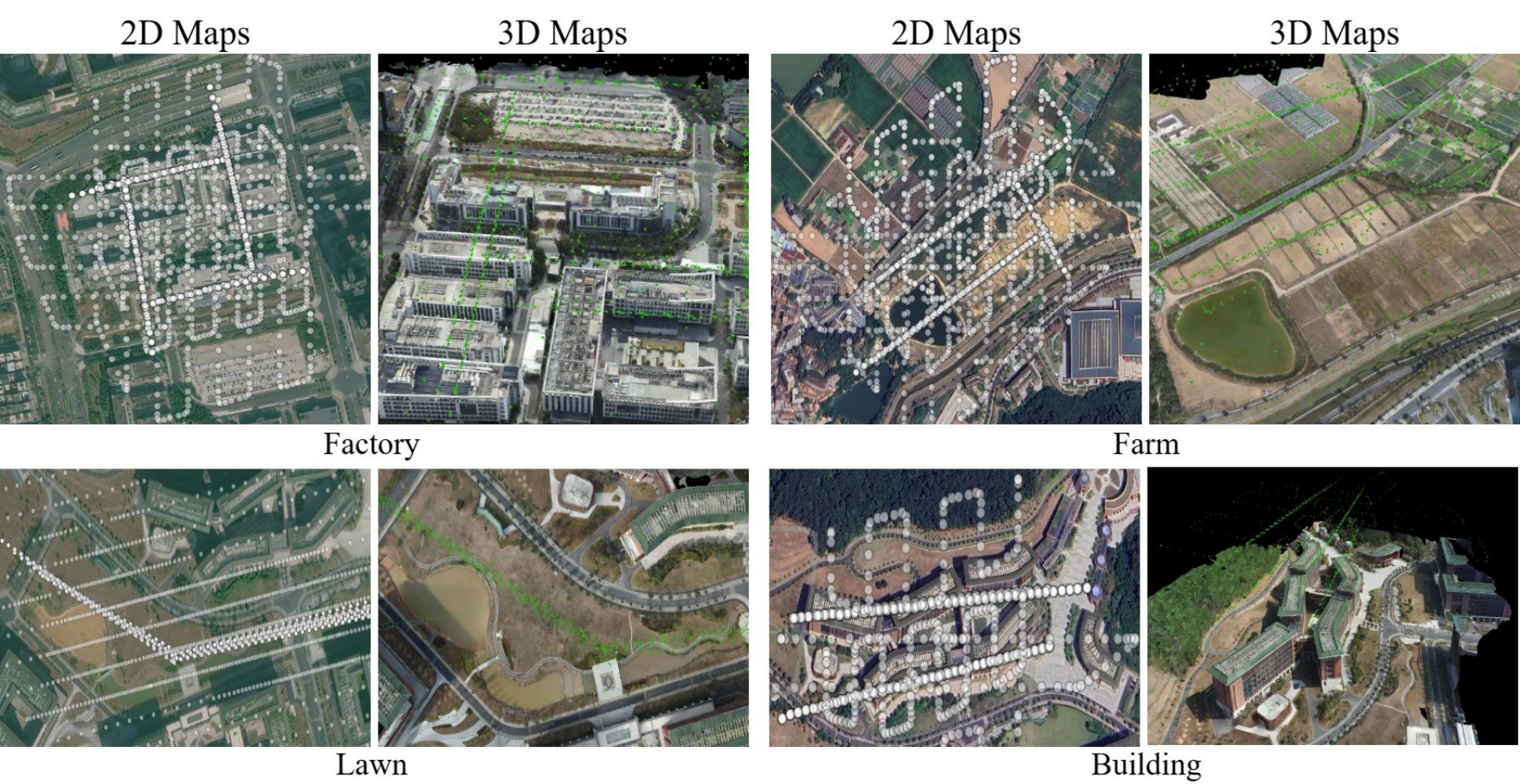}
    \caption{Automated flight trajectories for the \Decoupled dataset. The overlapping central segments represent the primary decoupled flight paths, while the peripheral trajectories correspond to auxiliary oblique photography missions designed to capture multi-angle perspectives for high-fidelity mesh reconstruction.}
    \label{fig:vis trail of decoupled}
\end{figure}

\subsection{\Decoupled: Variable Decoupled Dataset}

Standard benchmarks typically report aggregated performance metrics, which obscures the underlying causes of performance degradation for foundational depth models (e.g., ZoeDepth~\cite{bhat2023zoedepth}, MoGe2~\cite{wang2026moge}) in aerial domains. To systematically disentangle these domain gap factors, we introduce the \emph{\Decoupled} dataset. By providing an orthogonal parameter grid, this dataset enables the independent diagnosis of how Altitude, Pitch, and Field of View (FOV) individually impact metric depth estimation.

\begin{figure}[htbp]
    \centering
    \includegraphics[width=0.95\textwidth]{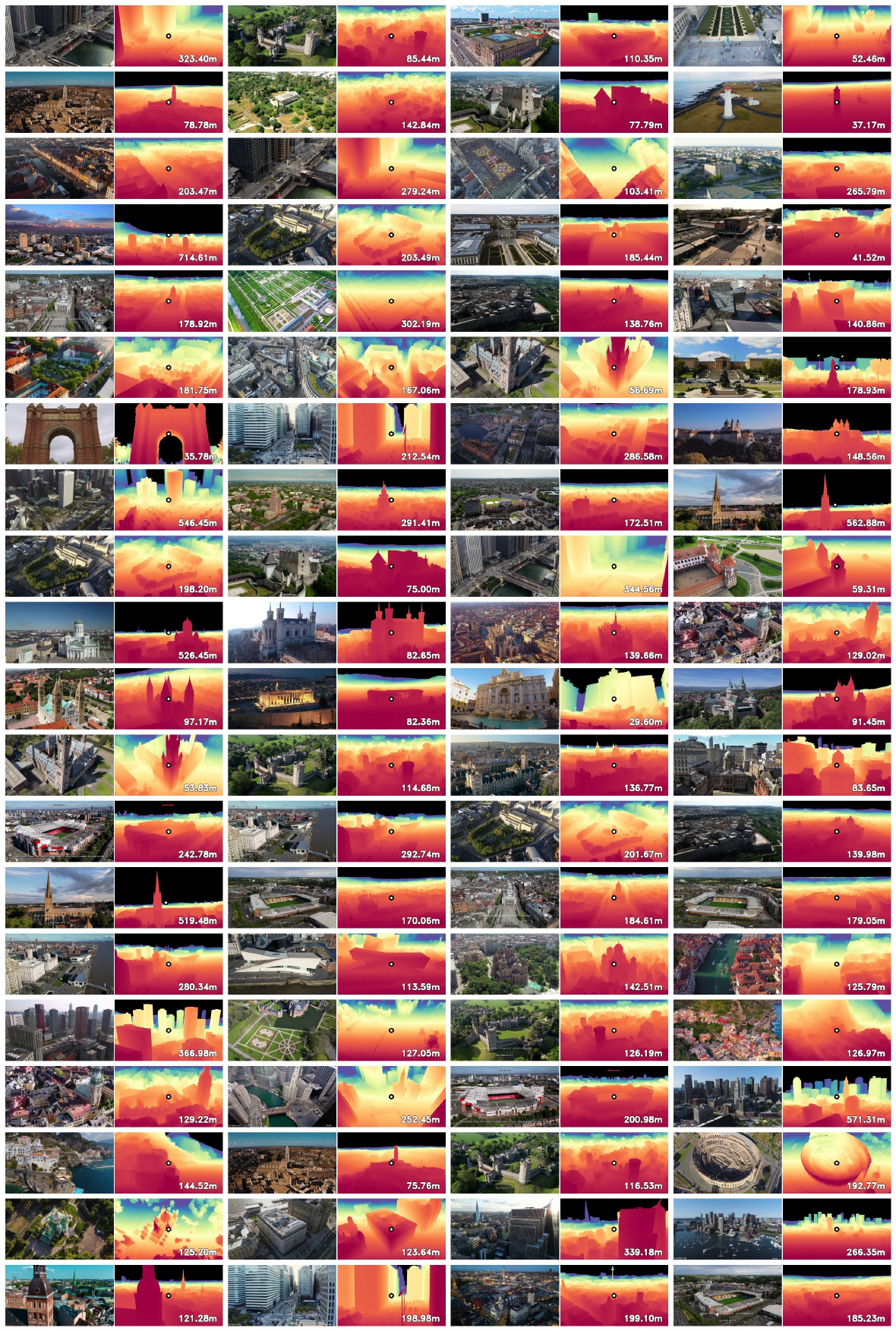} 
    \caption{Representative scenes from the \Wild dataset.}
    \label{fig:wild}
\end{figure}

The parameter grid covers typical operational ranges, including discrete altitudes ($h \in \{80, 120\}\text{m}$), pitch angles ($\theta \in \{-90^\circ, -75^\circ, -60^\circ, -45^\circ\}$), and two horizontal FOV settings ($83^\circ$ and $63^\circ$) to account for variations in scale and perspective.

This orthogonal sampling strategy results in a rich 3D matrix of $4 \times 2 \times 2 = 16$ distinct camera configurations for every single viewpoint in the \emph{\Decoupled} dataset. 

\paragraph{Automated Data Collection and Confounder Control.}
To decouple the effect of each spatial parameter, all imagery was captured via automated waypoint missions using DJI Pilot 2. \Fref{fig:vis trail of decoupled} visualizes the capturing trajectories. Each mission specifies 3D GPS waypoints with associated gimbal pitch angles and focal lengths. When varying one parameter, the remaining ones are held constant across all runs.

\subsection{\Wild: Reliable Pseudo-Metric Labels for Open-World Evaluation} 
\label{supp:sec_wild}

The \emph{\Wild} subset is curated from uncalibrated UAV videos sourced from the Internet and is designed to evaluate generalization under unconstrained in-the-wild conditions (see \fref{fig:wild} for scene visualizations). It should be interpreted as a pseudo-metric open-world evaluation set, rather than a sensor-level ground-truth benchmark. Starting from more than 600 raw clips, we retain 100 high-quality sequences after quality control for reconstruction reliability and usable scale references. These videos do not provide calibrated camera intrinsics or metric depth references, making metric depth recovery inherently ill-posed because of global scale and shift ambiguities. Rather than relying on COLMAP dense MVS, which can produce incomplete or noisy reconstructions for short, compressed, and irregular Internet UAV clips, we use COLMAP only for sparse pose and intrinsic recovery. We then employ DA3 for pose-conditioned multi-view dense inference and perform manual global scale calibration. This hybrid pipeline resolves the ambiguities in three stages and produces pseudo-metric depth labels. \begin{enumerate} \item \textbf{Sparse Pose Recovery and Up-to-Scale Geometry.} We first recover sparse camera poses and intrinsics using COLMAP~\cite{schonberger2016structure}. These estimates condition the DA3NESTED-GIANT-LARGE variant of Depth Anything 3 (DA3)~\cite{lin2025depth3} in multi-view mode. The resulting dense predictions are temporally consistent and defined up to a single sequence-level global scale. \item \textbf{Reference-Based Scale Measurement.} To resolve the remaining global scale, we collect physical reference lengths from each sequence. For distinctive scenes, we use VLMs (\eg, Gemini 2.5 Pro~\cite{comanici2025gemini}) to identify candidate architectural landmarks, which are manually verified and measured using Google Earth Pro~\cite{googleearthpro}. For generic scenes, 2--3 annotators independently measure multiple objects with known or standardized sizes, such as vehicles and lane markings. \item \textbf{Global Metric Alignment.} Given the physical length $L_{\text{real}}^{(i)}$ and the corresponding up-to-scale distance $L_{\text{uts}}^{(i)}$ for each reference object, we estimate a sequence-level global scale factor as $S = \frac{1}{N}\sum_{i=1}^{N} L_{\text{real}}^{(i)} / L_{\text{uts}}^{(i)}$, and obtain the final metric depth as $D_{\text{metric}} = S \times D_{\text{uts}}$. No per-frame scale fitting or access to evaluation ground truth is used during this process.
\end{enumerate}

\noindent
This multi-annotator, multi-object consensus ensures robust SfM-to-metric alignment.

\paragraph{Model Selection.} We use DA3 rather than MoGe2 for processing video sequences because MoGe2 does not natively enforce multi-view temporal consistency. To isolate the relative depth quality of the two inference methods, we evaluate them on held-out \emph{\Oblique} sequences with least-squares scale alignment. This alignment is used only for the controlled model-selection experiment and is not part of the \Wild pseudo-label generation pipeline. As shown in \Tref{tab:simplified_inference}, DA3 in multi-view mode achieves lower AbsRel and RMSE than MoGe2, while MoGe2 obtains a marginally higher $\delta_1$. We select DA3 because its multi-view inference provides more accurate metric geometry overall and better temporal consistency for video sequences.

\begin{table}[htbp]
    \centering
    \caption{Comparison of MoGe2 and DA3 under single-frame and multi-view inference, aligned using least squares.}
    \label{tab:simplified_inference}
    
    \renewcommand{\arraystretch}{1.2}
    \setlength{\tabcolsep}{14pt} %
    \small 
    
    \begin{tabular}{@{} l c c c c @{}}
        \toprule
        Model & Input & AbsRel $\downarrow$ & RMSE $\downarrow$ & $\delta_1 \uparrow$ \\
        \midrule
        
        MoGe2 & Single     & 3.25 & 5.55 & \textbf{99.98} \\ 
        \addlinespace
        
        DA3   & Single     & 4.57 & 7.25 & 99.79 \\
        DA3   & Multi-View & \textbf{1.18} & \textbf{2.69} & 99.97 \\
        
        \bottomrule
    \end{tabular}
\end{table}

\paragraph{Quantitative Error Boundary Analysis.} To quantify the reliability of the DA3-based pseudo-metric label generation pipeline, we conduct a proxy experiment on a held-out \emph{\Oblique} subset with high-quality MVS ground truth. We withhold the original calibrated camera metadata, treat the sequences as uncalibrated Internet videos, recover sparse poses and intrinsics with COLMAP, and run the complete DA3 plus manual scale-calibration pipeline. We then compare the generated pseudo-metric labels against the actual MVS ground truth. Importantly, no post-hoc depth alignment to the MVS ground truth is applied.

As reported in \Tref{tab:merged_inference_ablation2}, the resulting pseudo-metric labels achieve an AbsRel of 1.19\%, an RMSE of 2.66\,m, and a $\delta_1$ of 99.98\%. These results quantify the metric-scale error introduced by the pseudo-label generation process under a validated setting, supporting the use of \Wild for open-world generalization evaluation while not equating it with sensor-level ground truth.

\paragraph{Interpretation of \Wild Performance.} Compared with the controlled subsets, lower model performance on \Wild is consistent with its unconstrained viewpoints, heterogeneous video quality, and broader depth range; the latter is illustrated in Fig.~\ref{fig:depth_statistics}.

\begin{table}[htbp]
    \centering
    \caption{Unaligned metric error analysis between the pseudo-metric depth and the ground truth.}
    \label{tab:merged_inference_ablation2}
    
     \renewcommand{\arraystretch}{1.2}
    \setlength{\tabcolsep}{14pt} %
    \small 
    
    \begin{tabular}{@{} l c c c c @{}}
        \toprule
        Model & Input & AbsRel $\downarrow$ & RMSE $\downarrow$ & $\delta_1 \uparrow$ \\
        \midrule
        DA3   & Multi-View & \textbf{1.19} & \textbf{2.66} & 99.98 \\
        
        \bottomrule
    \end{tabular}
\end{table}

\subsection{\Synthetic: Variable Controlled Rendering}
\label{supp:sec_synthetic}

While \emph{\Oblique} and \emph{\Decoupled} provide crucial real-world domain fidelity, scaling them to cover continuous parameter distributions is prohibitively expensive. To bridge this gap, we introduce \emph{\Synthetic}, which offers scalable, cost-effective modulation of spatial parameters (\eg, altitude, pitch, FOV), enabling simulation of arbitrary UAV flight configurations. Note that \emph{\Synthetic} is used exclusively for training.

\begin{figure}[htbp]
    \centering
    \includegraphics[width=0.9\linewidth]{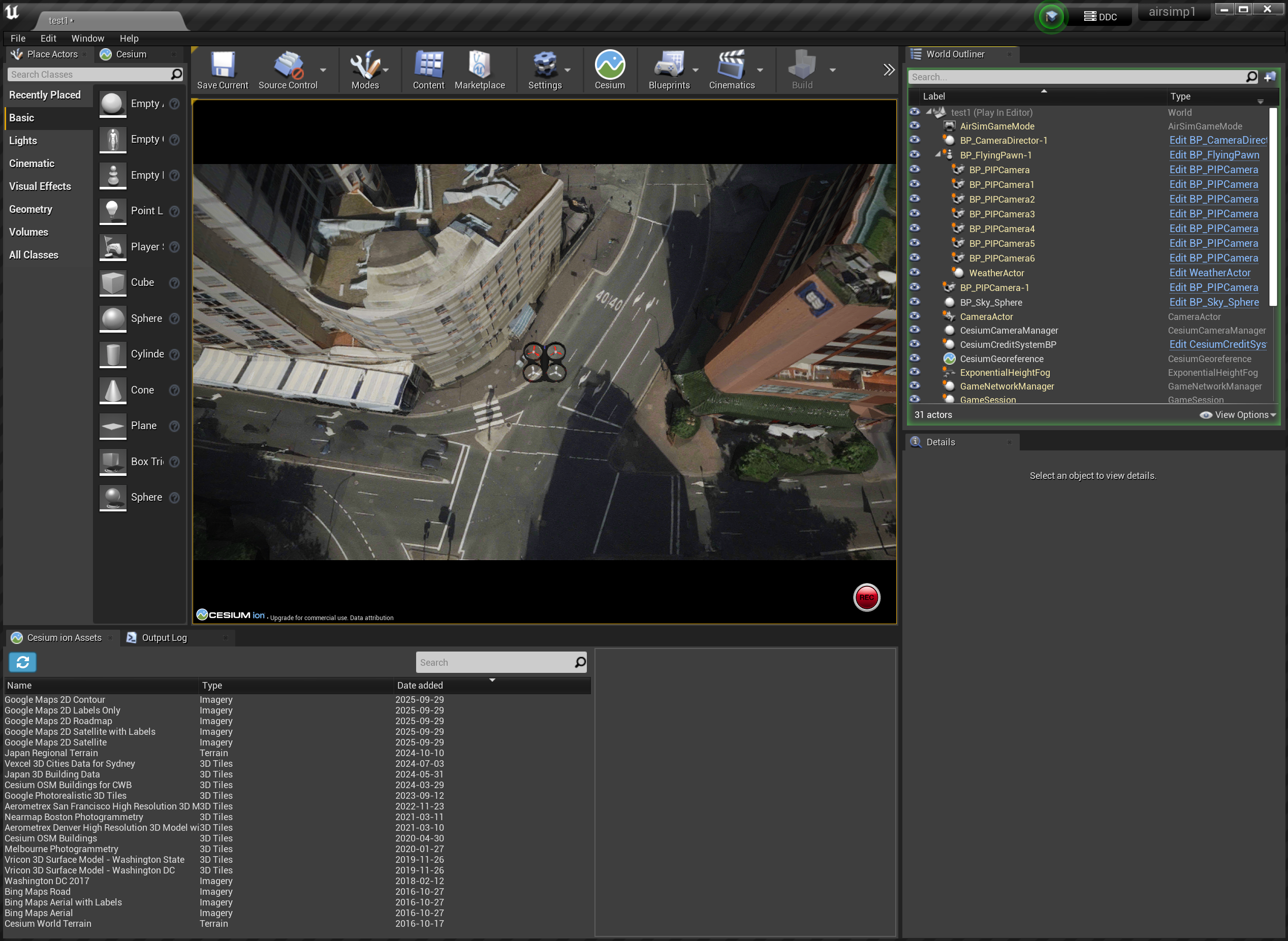}
    \caption{Photorealistic simulation environment in UE4. By streaming massive 3D tiles via Cesium and utilizing AirSim for UAV control, our pipeline facilitates the scalable generation of synthetic data with continuous, parameterized camera poses and accurate Z-buffer depth.}
    \label{fig:vis of ue}

\end{figure}

\paragraph{Photorealistic Simulation Pipeline.} To mitigate the sim-to-real domain gap, we build a high-fidelity simulation pipeline in Unreal Engine 4 (UE4) with the AirSim plugin for programmatic UAV control (\cref{fig:vis of ue}). We stream photogrammetry-based 3D assets, including Vexcel's \emph{Sydney} scene and Google 3D Tiles, via the Cesium plugin to ensure structural and visual realism. Through the AirSim Python API, we systematically vary spatial parameters (\ie, altitude, pitch, and FOV) along simulated flight trajectories, enabling scalable generation of strictly controlled training data with artifact-free, pixel-aligned metric depth from the Z-buffer.

\begin{figure}[htbp]
    \centering
    \includegraphics[width=\linewidth]{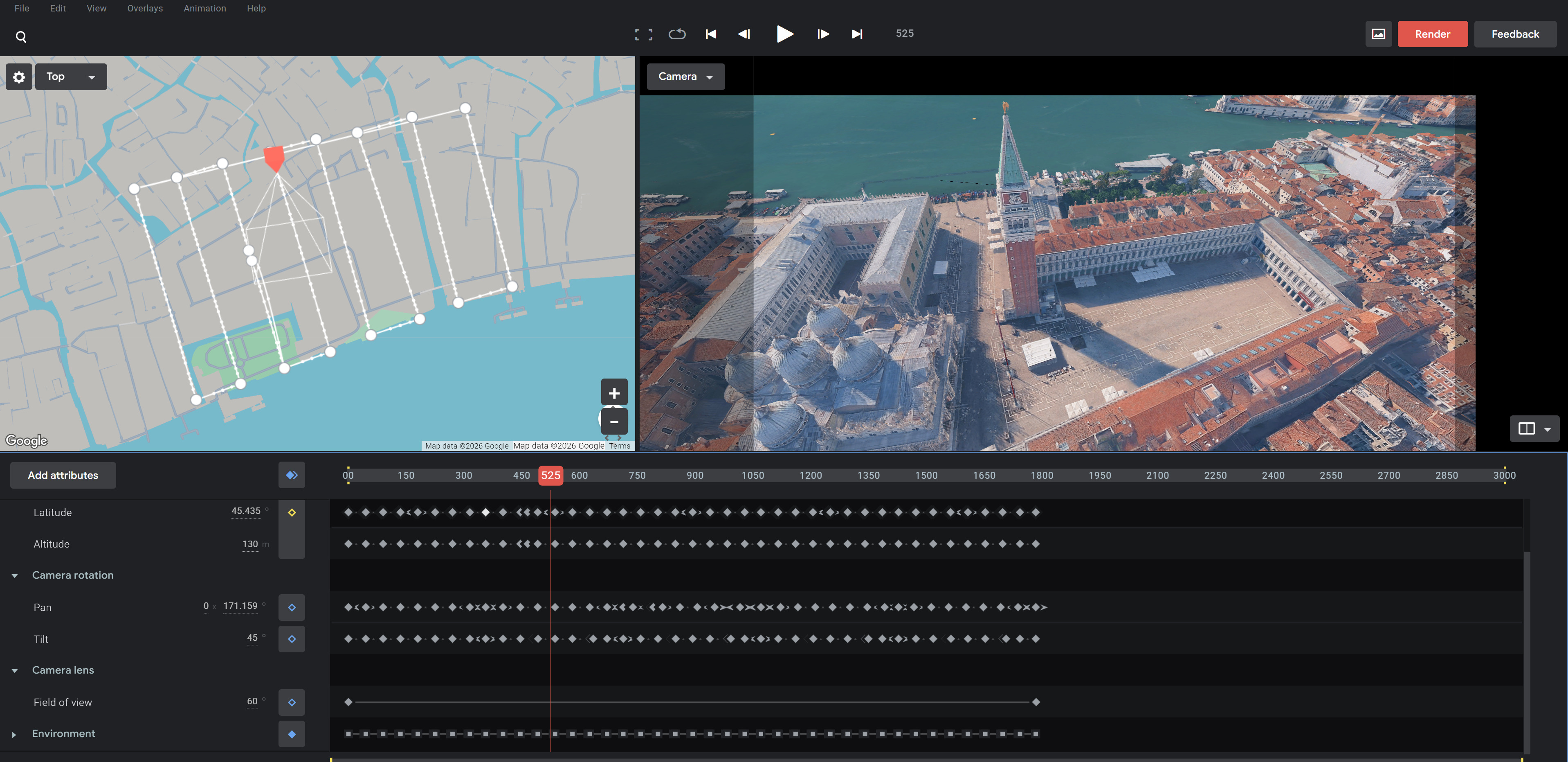}
    \caption{Google Earth Studio rendering pipeline for synthetic data generation.}
    \label{fig:googledearth}
\end{figure}

\paragraph{Google Earth Studio Pipeline.} While our UE4 simulation provides exact depth annotations, scaling it to encompass diverse global cityscapes is constrained by 3D asset availability and rendering overhead. To efficiently augment our synthetic data with varied urban architectures, we introduce a complementary pipeline using Google Earth Studio, as illustrated in \cref{fig:googledearth}. Its keyframe-based interface allows systematic manipulation of continuous spatial parameters (\ie, altitude, pitch, and FOV) to render ultra-high-resolution (3840$\times$2160) image sequences. Since GES lacks native dense depth exports, we process these sequences through the same pipeline used for our real-world \Oblique dataset, ensuring that the resulting depth maps share consistent structural characteristics and noise distributions with our empirical GT.

\section{More Details for the Model Finetuning}
In the main manuscript, we have presented the ablation results on fine-tuning strategies. Here, we provide a more detailed analysis of the fine-tuning configurations and design choices.

\subsection{Fine-tuning Implementation Details}

Since UAV images vary in resolution and aspect ratio, we adopt a dynamic token range strategy, which resizes each input image so that its longer side falls within a predefined range, yielding 1200--3600 visual tokens per image after patchification. We further apply data augmentation (\eg, color jittering, JPEG compression, Gaussian blur, and shot noise) to ensure robust feature learning.

\paragraph{Full-Parameter Fine-tuning.} 
For the full-parameter fine-tuning and the frozen-ViT variants, the models are trained for over $9,000$ steps using the AdamW optimizer. We employ a Sequential Learning Rate Scheduler with a $500$-step linear warmup followed by a cosine annealing decay down to $1 \times 10^{-6}$. In this full fine-tuning setting, we utilize the complete loss suite provided by the foundational architecture, encompassing global affine-invariant loss, metric scale loss, normal loss, mask loss, and local patch losses, to comprehensively update the entire feature representation.

\paragraph{Ineffectiveness of Scale-head only Fine-tuning.} 
By design, our foundation model, MoGe2\cite{wang2026moge}, inherently decouples metric scale estimation into a distinct \texttt{scale\_head}. A naive adaptation strategy might involve freezing the backbone and solely fine-tuning this scale head. However, we empirically observed that a head-only tuning strategy leads to highly unstable training and severe oscillation. 
This instability stems from the frozen backbone: having been trained on ground-level datasets, it imposes erroneous FOV and perspective priors when processing aerial images, resulting in severely distorted relative 3D point clouds. Consequently, the isolated scale head is forced to predict extreme, oscillating offsets to forcibly align a fundamentally incorrect relative geometry with the true aerial metric scale.
Therefore, a joint optimization strategy is necessary. 

\paragraph{LoRA Fine-tuning Configuration.} 
For the Low-Rank Adaptation (LoRA) experiments, instead of relying on the complete loss suite, we disable the dense local losses (i.e., normal loss, mask loss, and patch loss). 
The reason is that the pre-trained foundation model has already established robust high-frequency structural priors (e.g., sharp edges and generic object boundaries). By deactivating dense local losses, we prevent the limited capacity of the LoRA modules from overfitting to potential local noise or artifacts in the MVS ground truth, thereby avoiding the degradation of local sharpness (blurring). Instead, we focus the LoRA updates entirely on rectifying the domain-specific global geometry and metric scale of the UAV perspective. To this end, we only retain the global affine-invariant loss (weight $\lambda_{global} = 1.0$) and the metric scale loss (weight $\lambda_{metric} = 1.5$). 

For parameter-efficient fine-tuning, we inject LoRA modules into both the multi-head self-attention mechanisms (target modules: \texttt{qkv} and \texttt{proj}) and the feed-forward networks (target modules: \texttt{fc1} and \texttt{fc2}) within the ViT backbone. The LoRA hyperparameters are configured with a rank of $r=96$, an alpha of $\alpha=192$, and a dropout rate of $0.1$. 
During our preliminary experiments, we empirically observed that full-parameter fine-tuning achieved the highest quantitative performance on our \Oblique benchmark, which we attribute to the highly complex distributions of aerial altitudes, FOVs, and pitch angles. To approximate this upper-bound capacity while maintaining training efficiency, a relatively large rank ($r=96$) is strictly required to adapt to the severe geometric distortions and massive domain gaps of the UAV perspective.

\begin{table*}[htbp]
\centering
\caption{Evaluation under decoupled camera parameters on \Decoupled.}
\label{tab:parameter_decoupling}

\renewcommand{\arraystretch}{1.15}
\resizebox{\textwidth}{!}{
\begin{tabular}{c l c @{\hspace{1.2em}} r r r @{\hspace{1.2em}} r r r @{\hspace{1.2em}} r r r @{\hspace{1.2em}} r r r}
\toprule

\multirow{2}{*}{\textbf{Factor}} & \multirow{2}{*}{\textbf{Method}} & \multirow{2}{*}{\textbf{Setting}} &
\multicolumn{3}{c}{\textbf{Building}} & %
\multicolumn{3}{c}{\textbf{Factory}} &  %
\multicolumn{3}{c}{\textbf{Farm}} &     %
\multicolumn{3}{c}{\textbf{Lawn}} \\    %
\cmidrule(lr){4-6} \cmidrule(lr){7-9} \cmidrule(lr){10-12} \cmidrule(lr){13-15}

& & &
\multicolumn{1}{@{\hspace{1.2em}}c}{AbsRel} & \multicolumn{1}{c}{RMSE} & \multicolumn{1}{c}{$\delta_{1}$} &
\multicolumn{1}{@{\hspace{1.2em}}c}{AbsRel} & \multicolumn{1}{c}{RMSE} & \multicolumn{1}{c}{$\delta_{1}$} &
\multicolumn{1}{@{\hspace{1.2em}}c}{AbsRel} & \multicolumn{1}{c}{RMSE} & \multicolumn{1}{c}{$\delta_{1}$} &
\multicolumn{1}{@{\hspace{1.2em}}c}{AbsRel} & \multicolumn{1}{c}{RMSE} & \multicolumn{1}{c}{$\delta_{1}$} \\
\midrule

\multirow{6}{*}{\textbf{FOV}}
& \multirow{2}{*}{MoGe2 } & 63$^\circ$ & 27.92 & 30.63 & 30.6 & 31.35 & 36.77 & 24.4 & 74.01 & 96.82 & 0.1 & 35.40 & 43.70 & 17.3 \\
& & 83$^\circ$ & 30.01 & 35.00 & 22.6 & 22.40 & 26.01 & 44.8 & 67.40 & 87.11 & 0.0 & 29.96 & 36.08 & 21.4 \\
\cmidrule{2-15}
& \multirow{2}{*}{UniDepthV2} & 63$^\circ$ & 28.56 & 29.85 & 20.5 & 27.29 & 29.58 & 20.8 & \textbf{22.22} & 30.17 & 35.9 & 30.23 & 35.04 & 11.5 \\
&                             & 83$^\circ$ & 19.58 & 23.03 & 49.4 & 16.63 & 19.20 & 72.6 & \textbf{11.13} & \textbf{15.86} & \textbf{84.9} & 18.67 & 23.10 & 56.0 \\
\cmidrule{2-15}
& \multirow{2}{*}{MoGe2-Aerial} & 63$^\circ$ & \textbf{12.10} & \textbf{12.16} & \textbf{85.5} & \textbf{12.22} & \textbf{13.88} & \textbf{87.5} & 23.49 & \textbf{29.78} & \textbf{49.3} & \textbf{11.31} & \textbf{14.59} & \textbf{90.6} \\
& & 83$^\circ$ & \textbf{11.55} & \textbf{13.05} & \textbf{88.8} & \textbf{8.60} & \textbf{9.99} & \textbf{95.8} & 19.03 & 24.73 & 64.0 & \textbf{9.14} & \textbf{11.38} & \textbf{98.8} \\
\midrule

\multirow{12}{*}{\textbf{Pitch}}
& \multirow{4}{*}{MoGe2 } & -90$^\circ$  & 31.99 & 29.61 & 14.6 & 34.50 & 33.19 & 7.3 & 74.04 & 80.50 & 0.0 & 40.56 & 40.91 & 4.6 \\
& & -75$^\circ$ & 24.03 & 23.33 & 37.3 & 21.13 & 21.83 & 47.1 & 68.77 & 78.24 & 0.0 & 29.78 & 31.82 & 26.2 \\
& & -60$^\circ$ & 22.90 & 25.84 & 42.8 & 18.10 & 21.39 & 63.8 & 66.87 & 86.48 & 0.0 & 25.68 & 31.40 & 34.1 \\
& & -45$^\circ$  & 36.46 & 51.49 & 12.7 & 33.56 & 49.66 & 21.2 & 74.51 & 124.55 & 0.1 & 35.12 & 54.18 & 12.1 \\
\cmidrule{2-15}
& \multirow{4}{*}{UniDepthV2} & -90$^\circ$ & 27.85 & 25.93 & 25.0 & 27.23 & 26.44 & 26.0 & \textbf{15.43} & \textbf{17.79} & \textbf{65.8} & 26.86 & 27.01 & 28.5 \\
&                             & -75$^\circ$ & 23.78 & 23.06 & 36.1 & 21.72 & 21.81 & 48.0 & \textbf{13.87} & \textbf{16.58} & \textbf{69.3} & 24.17 & 25.25 & 36.8 \\
&                             & -60$^\circ$ & 20.91 & 23.40 & 45.2 & 18.54 & 20.83 & 61.5 & \textbf{18.37} & \textbf{24.61} & \textbf{51.4} & 24.47 & 28.92 & 35.4 \\
&                             & -45$^\circ$ & 23.70 & 33.02 & 33.8 & 19.97 & 28.49 & 52.9 & 21.30 & 36.00 & 45.0 & 22.59 & 34.37 & 34.0 \\
\cmidrule{2-15}
& \multirow{4}{*}{MoGe2-Aerial} & -90$^\circ$  & \textbf{13.50} & \textbf{13.00} & \textbf{78.8} & \textbf{10.92} & \textbf{10.65} & \textbf{90.0} & 19.97 & 21.04 & 60.7 & \textbf{13.00} & \textbf{13.72} & \textbf{90.4} \\
& & -75$^\circ$ & \textbf{11.10} & \textbf{10.82} & \textbf{87.4} & \textbf{10.74} & \textbf{10.72} & \textbf{90.8} & 21.80 & 24.97 & 55.9 & \textbf{10.10} & \textbf{11.43} & \textbf{95.9} \\
& & -60$^\circ$ & \textbf{12.46} & \textbf{12.38} & \textbf{89.4} & \textbf{11.08} & \textbf{12.65} & \textbf{90.4} & 24.43 & 31.45 & 47.3 & \textbf{9.04} & \textbf{11.72} & \textbf{95.9} \\
& & -45$^\circ$  & \textbf{10.29} & \textbf{14.11} & \textbf{92.8} & \textbf{8.81} & \textbf{13.88} & \textbf{95.6} & \textbf{19.75} & \textbf{32.58} & \textbf{59.8} & \textbf{9.08} & \textbf{14.94} & \textbf{96.2} \\
\midrule

\multirow{6}{*}{\textbf{Altitude}}
& \multirow{2}{*}{MoGe2 } & 80m & 24.04 & 20.58 & 43.1 & 22.74 & 20.12 & 47.1 & 68.15 & 70.27 & 0.1 & 29.81 & 29.38 & 30.2 \\
& & 120m & 33.89 & 45.05 & 10.1 & 31.01 & 42.66 & 22.0 & 73.41 & 110.58 & 0.0 & 35.55 & 50.42 & 8.5 \\
\cmidrule{2-15}
& \multirow{2}{*}{UniDepthV2} & 80m  & 20.28 & 17.09 & 52.5 & 19.84 & 16.96 & 56.9 & \textbf{12.25} & \textbf{12.97} & \textbf{77.9} & 21.80 & 20.31 & 48.3 \\
&                             & 120m & 27.86 & 35.80 & 17.4 & 24.08 & 31.82 & 36.6 & 21.33 & 32.56 & 41.5 & 27.08 & 37.83 & 19.2 \\
\cmidrule{2-15}
& \multirow{2}{*}{MoGe2-Aerial} & 80m & \textbf{11.79} & \textbf{9.54} & \textbf{86.6} & \textbf{12.18} & \textbf{11.47} & \textbf{88.0} & 23.51 & 24.43 & 54.8 & \textbf{10.89} & \textbf{12.10} & \textbf{92.2} \\
& & 120m & \textbf{11.86} & \textbf{15.67} & \textbf{87.8} & \textbf{8.63} & \textbf{12.39} & \textbf{95.3} & \textbf{19.84} & \textbf{30.04} & \textbf{56.8} & \textbf{9.55} & \textbf{13.86} & \textbf{97.2} \\

\bottomrule
\end{tabular}
}

    \centering
    \includegraphics[width=0.95\textwidth]{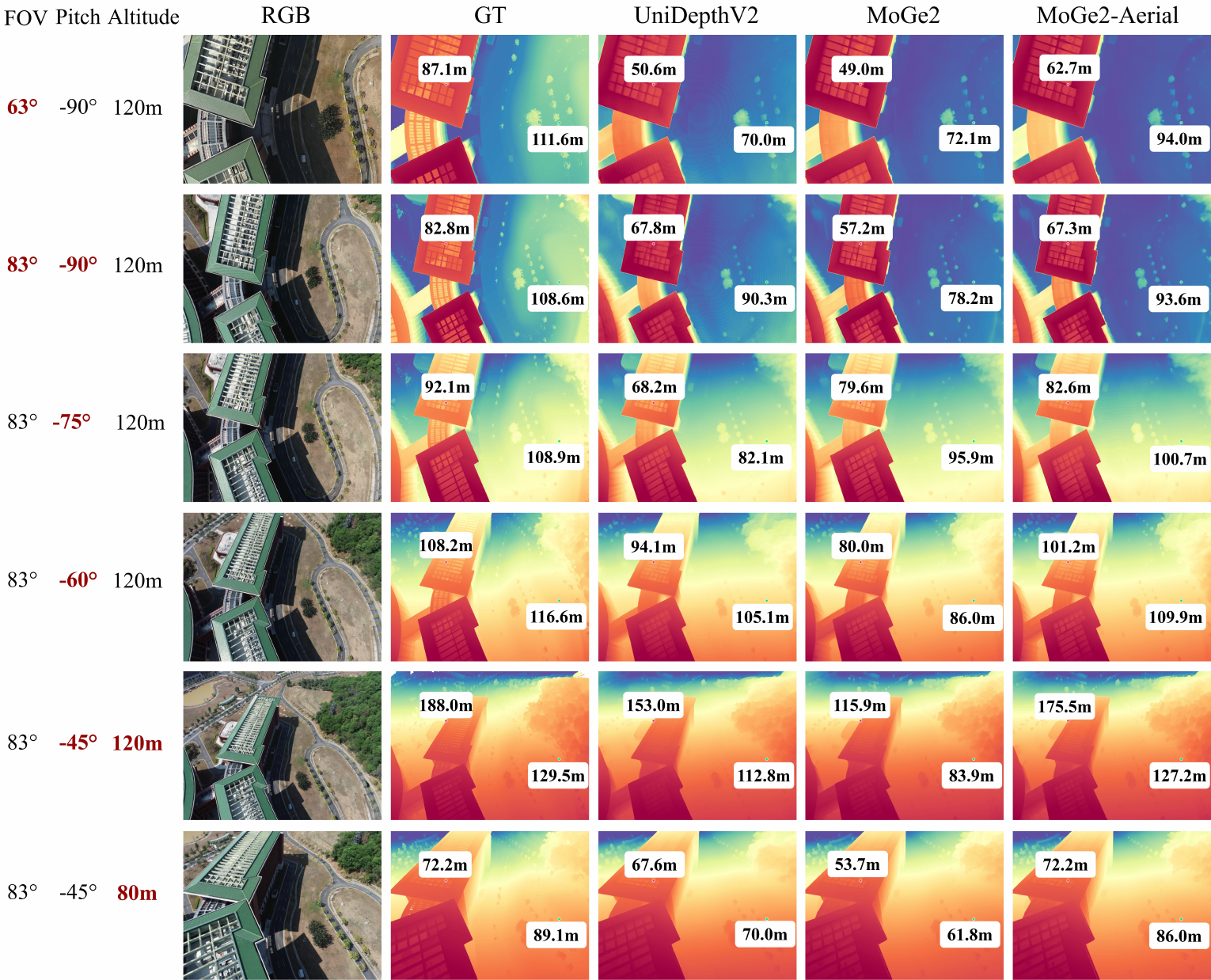}
    \captionof{figure}{Results on \Decoupled across FOV, pitch, and altitude.}
    \label{fig:2dmaps of fov}

\end{table*}

\section{More Experimental Results}
\label{supp:sec_more_results}

\subsection{Detailed Evaluation on the \Decoupled dataset}
A key advantage of our \Decoupled benchmark is its orthogonal parameter design, where altitude, pitch, and FOV are independently varied while the others remain fixed. This allows us to isolate the individual impact of each parameter on depth estimation.
\Tref{tab:parameter_decoupling} presents the quantitative results under each decoupled setting, from which several findings emerge. First, FOV variation introduces non-negligible degradation. Furthermore, pitch angle substantially affects estimation quality, particularly at extreme nadir views (Pitch $\approx -90^\circ$), where the lack of horizon cues causes significant performance drops in baselines, while our model is largely unaffected. Similarly, altitude is a significant contributing factor to depth estimation degradation. Baseline models exhibit severe scale drift as altitude increases to 120m, whereas our MoGe2-Aerial remains stable across the full altitude range. These decoupled analyses, qualitatively corroborated by the depth maps in \cref{fig:2dmaps of fov}, confirm that altitude, viewing angle, and FOV each independently pose substantial challenges for existing monocular depth estimators in aerial scenarios. By systematically covering these parameter dimensions, our \emph{AerialMetric} benchmark fills a critical gap left by previous aerial depth datasets.

\subsection{More Qualitative Evaluations}
Due to the page limits of the main manuscript, we present additional qualitative comparisons to further demonstrate the effectiveness of our MoGe2-Aerial against the state-of-the-art methods, including DepthPro~\cite{Bochkovskii2024:arxiv}, ZoeDepth~\cite{bhat2023zoedepth}, Uni-Depth V2~\cite{piccinelli2025unidepthv2}, and MoGe2~\cite{wang2026moge}.

\begin{figure*}[b] %
    \centering
    \includegraphics[width=\textwidth]{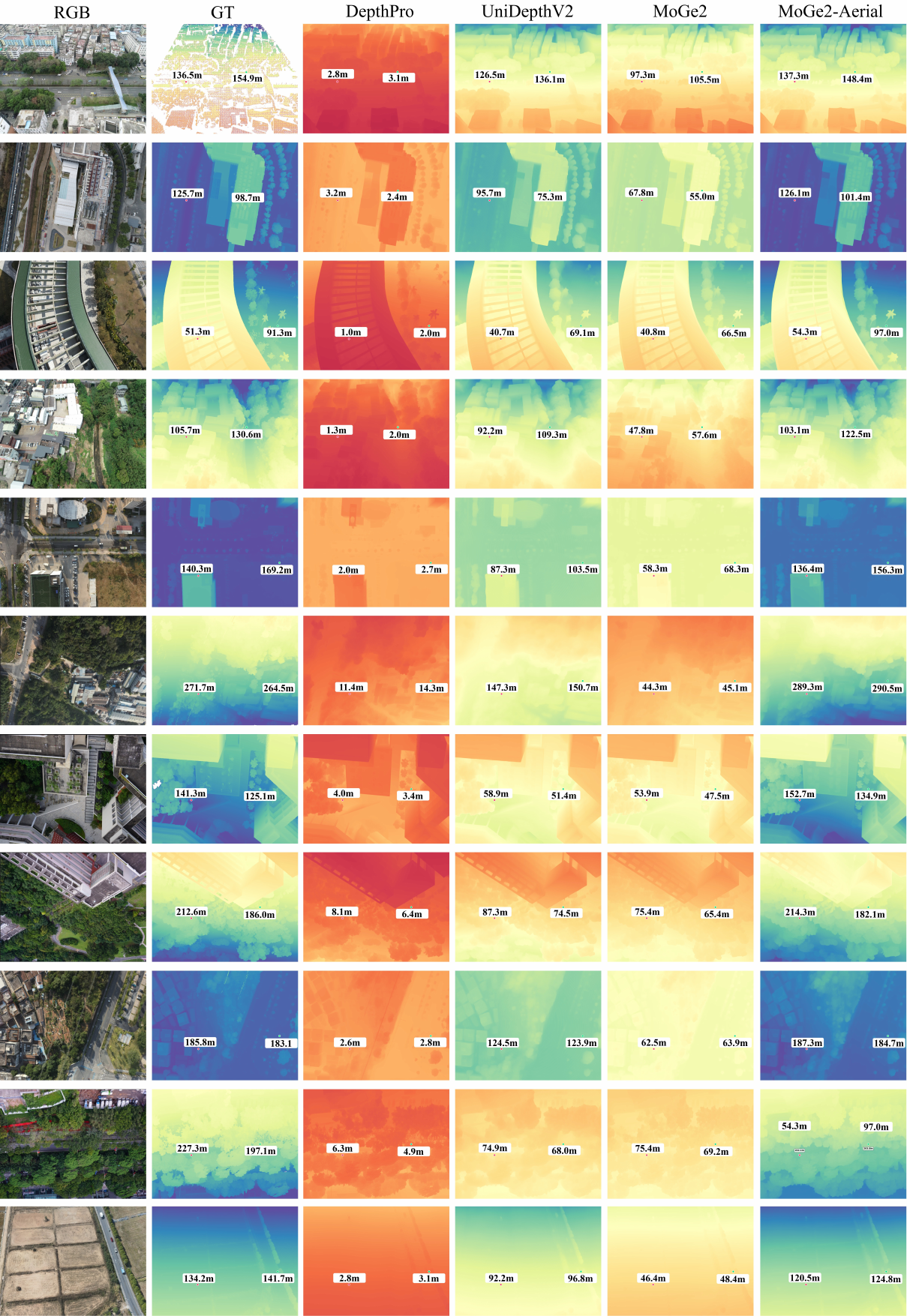}
    \caption{Comprehensive qualitative comparison of 2D metric depth maps.}
    \label{fig:supp_qualitative_2d_1}
\end{figure*}

In \cref{fig:supp_qualitative_2d_1}, we present high-resolution depth predictions across diverse real environments. Baseline models frequently suffer from scale ambiguity, yielding erroneous absolute distances, and fail to preserve fine structural details, resulting in blurred object boundaries. In contrast, MoGe2-Aerial maintains sharp edge delineation and highly accurate metric ranges strictly aligned with the ground truth.

\begin{figure*}[tbp]
    \centering
    \includegraphics[width=\textwidth]{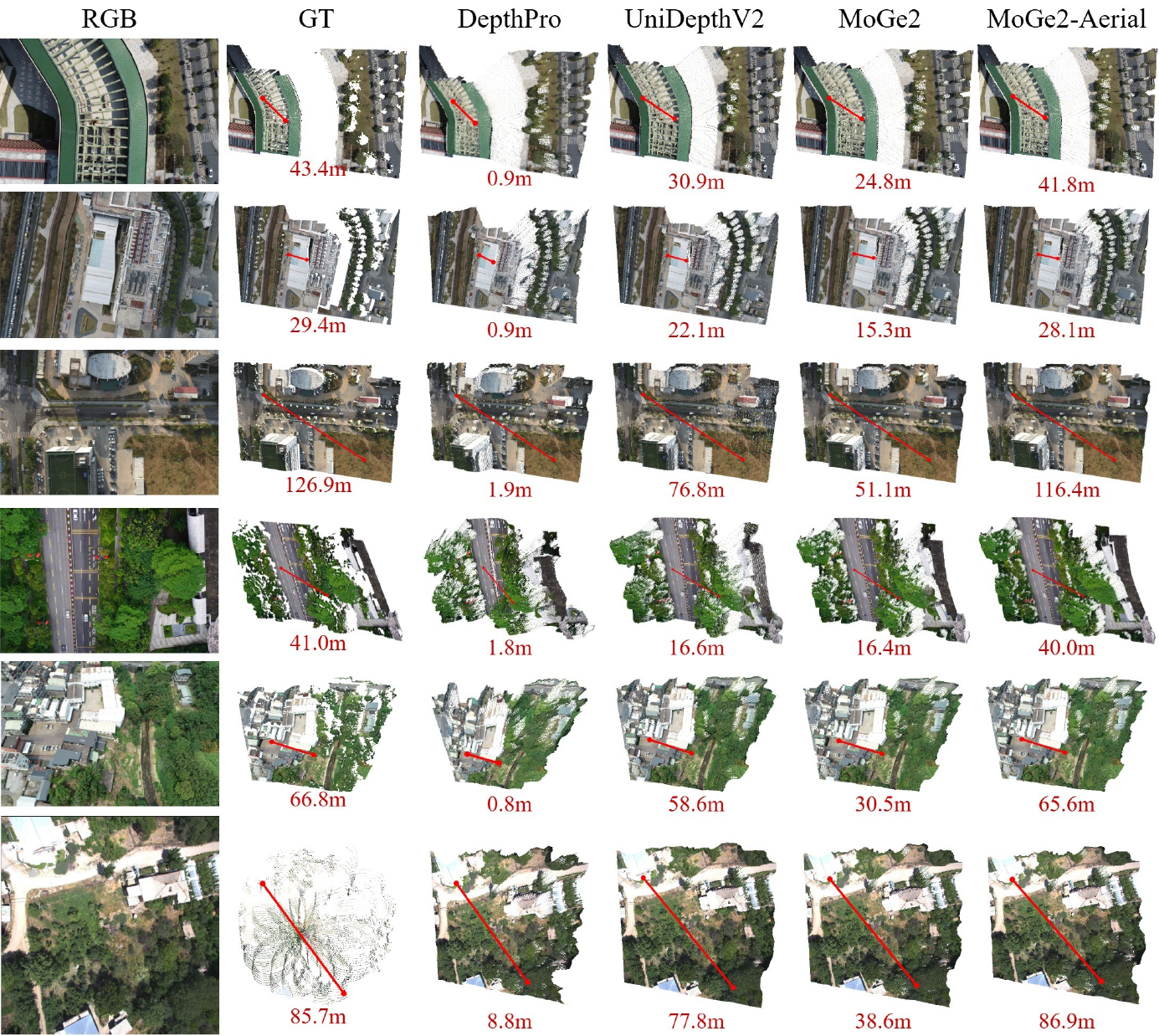}
    \caption{3D point cloud comparison under UAV perspective.}
    \label{fig:supp_qualitative_3d}
\end{figure*}

Furthermore, we unproject the predicted depth maps into 3D point clouds using the corresponding camera intrinsics. As illustrated in \cref{fig:supp_qualitative_3d}, baseline models suffer from severe structural warping due to incorrect FOV priors, manifesting as geometric distortions such as sloped façades and curved ground planes. Conversely, our model successfully preserves the structural integrity of the scenes, recovering vertical façades and flat ground planes, which demonstrates a robust geometric understanding of complex aerial topographies.

\begin{table}[!t]
\centering
\caption{Cross-architecture adaptation results on AerialMetric.}
\label{tab:aerial_mean}
\setlength{\tabcolsep}{1.8pt}
\renewcommand{\arraystretch}{0.88}
\scriptsize
\resizebox{\linewidth}{!}{
\begin{tabular}{lccccccccc}
\toprule
\multirow{2}{*}{Method}
& \multicolumn{4}{c}{\textbf{AerialMetric-Oblique}}
& \multicolumn{4}{c}{\textbf{AerialMetric-Decoupled}}
& \multicolumn{1}{c}{\textbf{Farm}} \\
\cmidrule(lr){2-5} \cmidrule(lr){6-9} \cmidrule(lr){10-10}
& AbsRel $\downarrow$ & RMSE $\downarrow$ & $\delta_1 \uparrow$ & $\overline{\delta_{1}} \uparrow$
& AbsRel $\downarrow$ & RMSE $\downarrow$ & $\delta_1 \uparrow$ & $\overline{\delta_{1}} \uparrow$
& $\delta_1 \uparrow$ \\
\midrule

\rowcolor{gray!10}
\multicolumn{10}{c}{\scriptsize{w/o GT intrinsics}} \\

UniDepthV2
& 28.6 & 45.15 & 39.1 & 2.39
& 22.4 & 26.52 & 41.6 & 2.37
& 57.9 \\

UniDepthV2-Aerial
& 21.4 & 31.37 & 57.8 & 5.22
& 14.6 & 16.77 & 72.3 & 11.03
& \textbf{81.1} \\

MoGe2
& 49.5 & 72.15 & 7.9 & 3.40
& 38.4 & 47.00 & 20.0 & 6.40
& 0.0 \\

MoGe2-Aerial
& \textbf{12.4} & \textbf{17.49} & \textbf{84.4} & \textbf{52.90}
& \textbf{13.2} & \textbf{15.49} & \textbf{83.9} & \textbf{47.50}
& 54.2 \\

\midrule

\rowcolor{gray!10}
\multicolumn{10}{c}{\scriptsize{w/ GT intrinsics}} \\

UniDepthV2
& 25.9 & 41.78 & 45.1 & 3.51
& 17.3 & 21.05 & 62.4 & 3.39
& 91.0 \\

UniDepthV2-Aerial
& 18.6 & 27.63 & 65.1 & 6.79
& \textbf{9.4} & \textbf{11.33} & \textbf{92.8} & 17.13
& \textbf{92.9} \\

MoGe2
& 38.8 & 58.29 & 17.7 & 6.60
& 26.2 & 32.65 & 53.0 & 25.10
& 0.8 \\

MoGe2-Aerial
& \textbf{10.5} & \textbf{15.19} & \textbf{89.0} & \textbf{58.5}
& 13.2 & 14.95 & 85.6 & \textbf{46.00}
& 71.2 \\

\bottomrule
\end{tabular}
}
\end{table}
\subsection{Zero-shot Generalization on \Wild Data}
\begin{figure*}[htbp]
    \centering
    \includegraphics[width=\textwidth]{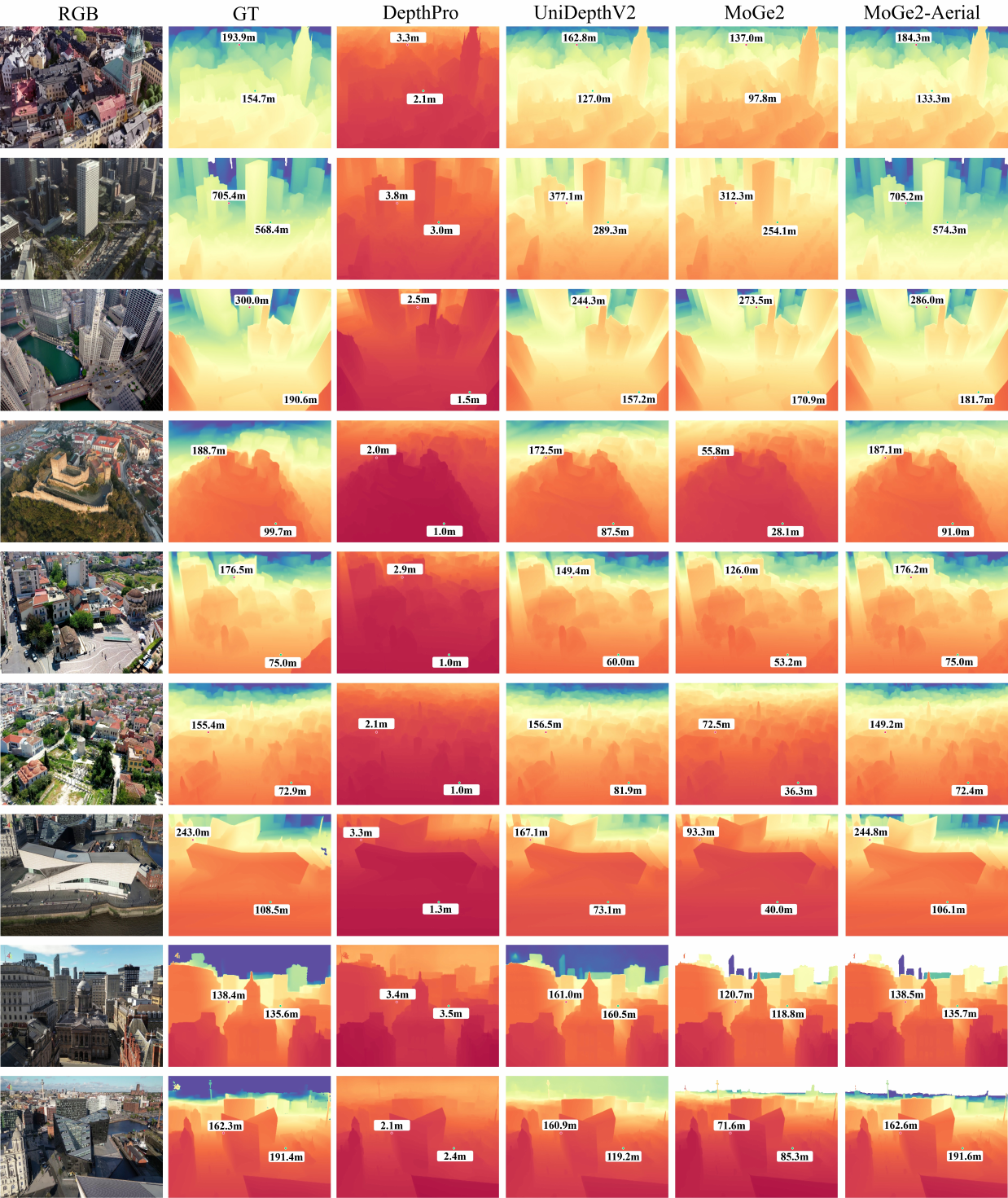}
    \caption{Zero-shot generalization on unseen \Wild videos. }
    \label{fig:supp_wild_generalization}
\end{figure*}

Our proposed \Wild dataset is reserved exclusively for evaluation, with none of its data used during training. This allows us to rigorously assess the zero-shot generalization capability of our model on completely unconstrained Internet UAV videos. As shown in \cref{fig:supp_wild_generalization}, we compare the predictions of MoGe2-Aerial against the state-of-the-art methods on these challenging in-the-wild scenes. The visualization results demonstrate that our model achieves accurate metric depth estimation across diverse and complex scenes, validating the strong generalization capability of our approach.

\subsection{Cross-Architecture Adaptation}

To further validate the effectiveness of AerialMetric beyond MoGe2\cite{wang2026moge}, we evaluate its transferability on UniDepthV2\cite{piccinelli2025unidepthv2}, a model with a substantially different architecture and training formulation. Since the official fine-tuning code of UniDepthV2\cite{piccinelli2025unidepthv2} is unavailable, we implement an adaptation pipeline for the released UniDepthV2-ViT-L/14 model and fine-tune it using LoRA on AerialMetric.

As shown in Table~\ref{tab:aerial_mean}, UniDepthV2-Aerial consistently improves over the original UniDepthV2\cite{piccinelli2025unidepthv2} on both the Oblique and Decoupled subsets, under settings with and without ground-truth camera intrinsics. These gains indicate that AerialMetric provides effective supervision for adapting monocular metric depth estimation models to UAV viewpoints.

Notably, UniDepthV2-Aerial achieves a $\delta_1$ score of 81.1 on the texture-homog-\\eneous Farm subset without ground-truth intrinsics, surpassing MoGe2 under the same setting. This result further demonstrates that the improvement is not limited to a single backbone, and that AerialMetric can provide transferable supervision across different depth estimation architectures.

Following TanDepth\cite{florea2025tandepth}, we additionally report the stricter threshold accuracy
$\overline{\delta_1}$, computed under tighter relative-error thresholds, to
better distinguish high-performing methods when the conventional
$\delta_1$ metric becomes saturated.

\end{document}